\documentclass{svproc}

\usepackage{url}

\usepackage[nolist]{acronym} 
\usepackage{graphicx}
\usepackage{bm}
\usepackage{amsmath}
\usepackage{amsfonts}
\usepackage{subcaption}
\usepackage{floatrow}
\usepackage{adjustbox}
\usepackage{url}
\usepackage[hidelinks]{hyperref}
\usepackage{gensymb}

\usepackage{epsfig}
\usepackage{pst-all, graphics, graphicx, color}
\usepackage[crop=pdfcrop]{pstool}
\usepackage{psfrag}
\usepackage{amsmath, amssymb}
\usepackage{epstopdf}
\usepackage{tikz}
\usepackage{multirow}
\usepackage{algorithm,algorithmicx,algpseudocode}
\usepackage{dsfont}
\usepackage{booktabs} 

\newfloatcommand{capbtabbox}{table}[][\FBwidth]

\newacro{vs}[VS]{Visual Servoing}
\newacro{il}[IL]{Imitation Learning}
\newacro{ds}[DS]{Dynamical System}
\newacro{dof}[DoF]{Degree of Freedom}     
\newacro{pbd}[PbD]{Programming by Demonstration}
\newacro{lfd}[LfD]{Learning from Demonstration}
\newacro{dmp}[DMP]{Dynamic Movement Primitive}
\newacro{rds}[RDS]{Reshaped Dynamical System}
\newacro{gmm}[GMM]{Gaussian Mixture Model}
\newacro{gmr}[GMR]{Gaussian Mixture Regression}
\newacro{rmse}[RMSE]{Root-Mean-Square Error}
\newacro{ibvs}[IBVS]{Image-based Visual Servoing}
\newacro{ilvs}[ILVS]{Imitation Learning for Visual Servoing}
\newacro{em}[EM]{Expectation-Maximization}

\begin{document}
\mainmatter              
\title{Imitation Learning-based Visual Servoing\\for Tracking Moving Objects}
\titlerunning{Imitation Learning-based Visual Tracking}  
%
\author{Rocco Felici\inst{1} \and Matteo Saveriano\inst{2} \and Loris Roveda\inst{3} \and Antonio Paolillo\inst{3}}
\authorrunning{R. Felici et al.} 
%
%
\institute{Università della Svizzera italiana (USI), Lugano, Switzerland \\
\email{rocco.felici@usi.ch},\\
\and
Department of Industrial Engineering (DII), \\ University of Trento, Trento, Italy 
\and
Dalle Molle Institute for Artificial Intelligence (IDSIA), \\ USI-SUPSI, Lugano, Switzerland}

\maketitle              

\begin{abstract}
In everyday life collaboration tasks between human operators and robots, the former necessitate simple ways for programming new skills, the latter have to show adaptive capabilities to cope with environmental changes. The joint use of visual servoing and imitation learning allows us to pursue the objective of realizing friendly robotic interfaces that (i) are able to adapt to the environment thanks to the use of visual perception and (ii) avoid explicit programming thanks to the emulation of previous demonstrations. 
This work aims to exploit imitation learning for the visual servoing paradigm to address the specific problem of tracking moving objects. 
In particular, we show that it is possible to infer from data the compensation term required for realizing the tracking controller, avoiding the explicit implementation of estimators or observers. 
The effectiveness of the proposed method has been validated through simulations with a robotic manipulator.

\end{abstract}

\section{Introduction}

%
Today robots are not merely asked to execute tasks in controlled environments, but they must have friendly interfaces so that everyone can conveniently operate them in everyday life.
In fact, given their high level of ubiquity, more and more robots are at the disposal of people with no technical expertise.
%
%
%
As a consequence, easy control frameworks that do not require specific engineering or programming skills are urgently needed. 
Furthermore, modern robots operating ``in the wild'' need to be highly adaptive, to cope with changes of dynamic environments.
%
%

%
\ac{il}~\cite{schaal1996learning}, also known as programming by demonstrations~\cite{billard2016learning} or learning from demonstrations~\cite{argall2009survey}, promises to avoid specific coding duties by imitating the desired behavior as performed by an expert~\cite{Caccavale2017Imitation}. 
%
With respect to classic control paradigms, \ac{il} is easier and more convenient for non-expert operators, as they only need to provide demonstrations of the desired robotic tasks.
%
%
Among the \ac{il} approaches, \ac{ds}-based methods~\cite{khansari2011learning,saveriano2020energy,saveriano2018incremental} allow realizing the imitation strategy while ensuring stability properties. 
Adaptive capabilities, instead, can be realized by including exteroceptive sensing, such as vision, into the \ac{il} strategy. 
In particular, recent work~\cite{paolillo2023dynamical,paolillo2022learning,valassakis2022demonstrate} have explored the possibility to combine \ac{vs}~\cite{chaumette2006visual,chaumette2007visual} with \ac{ds}-based \ac{il}. 
We name such integration \ac{ilvs}.
Such combination brings benefit to both techniques: on the one side, the visual perception adds adaptability to the \ac{il} scheme to cope with environmental changes; on the other, the imitation strategy allows the addition of tasks or constraints to the \ac{vs} law with no specific implementation.

This work aims at resorting to the \ac{ilvs} paradigm to tackle the specific problem of tracking moving objects. 
%
%
Traditional tracking techniques need to estimate the motion of the target, e.g., specifically implementing a Kalman filter~\cite{chaumette1993tracking} or predictive controllers~\cite{ginhoux2005active}.
Instead, we provide a framework that leverages \ac{ilvs} and extrapolates from demonstrations of tracking experiments the required information for adding the tracking skill to the basic \ac{vs} law. 
In particular, we propose to use the so-called \ac{rds} approach~\cite{saveriano2018incremental} to imitate the tracking behavior into the basic \ac{vs} control.
The resulting learning-aided control system has been validated with robotic simulations. 

\section{Background}\label{sec:background}


The well-known \ac{vs} technique~\cite{chaumette2006visual,chaumette2007visual} employs vision to control the motion of a robot. 
In particular, in image-based \ac{vs}, considered in this work, the objective is to zero the difference between desired and measured visual features that are directly defined on the camera image.
Such visual features represent the feedback of the controller that computes camera velocities to achieve a desired task; they can be detected with standard image processing~\cite{Marchand:ras:2005} or more sophisticated methods, e.g., artificial neural network~\cite{Paolillo:iros:2022}.
%
Assuming an eye-in-hand configuration, a static target, and constant desired features, the basic \ac{vs} law computes the camera velocity $\bm{v} \in \mathbb{R}^6$ with a simple reactive controller.
Its objective is to nullify the visual error $\bm{e}\in\mathbb{R}^k$ between the detected and desired visual features:
\begin{equation}
    \bm{v} = -\lambda \widehat{\bm{L}^+}\bm{e},
    \label{eq:vs}
\end{equation}
where $\lambda$ is a positive scalar gain and $\widehat{\bm{L}^+}\in \mathbb{R}^{6\times k}$ an approximation of the Moore-Penrose pseudoinverse of the interaction matrix~\cite{chaumette2006visual}.
Such approximation is normally due to unknown information, such as the depth of the visual features\footnote{To keep the notation compact, we omit the dependence of the interaction matrix on the visual features and their depth.}.
The simple law~\eqref{eq:vs} can be augmented with other tasks or constraints to enable additional skills, by employing planning techniques~\cite{Chesi:tro:2007,Mezouar:tro:2002}, predictive controllers~\cite{Allibert:tro:2010,Paolillo:icra:2020,Paolillo:cep:2023,Sauvee:cdc:2006}, and other sort of optimization-based frameworks~\cite{Agravante:ral:2017,Mingo:icra:2021,Paolillo:ral:2018}.
However, such approaches require careful design and implementation of the additional modules, which is desirable to avoid for the sake of easiness of use.

To this end, inspired by the \ac{ds} paradigm, it has been proposed to augment the skills of the basic law with an \ac{ilvs} strategy~\cite{paolillo2022learning}.
In particular, by using the specific \ac{rds} method~\cite{saveriano2018incremental}, one could write the augmented \ac{vs} law as follows:
\begin{equation}
    \bm{v}=-\lambda\widehat{\bm{L}^+} \bm{e}+h\bm{\rho}(\bm{e}),
    \label{eq:ilvs}
\end{equation}
where $\bm{\rho}(\bm{e})$ is an error-dependent corrective input used to follow complex trajectories and $h$ is a vanishing term used to suppress $\bm{\rho}$ after a user-defined time and retrieve stability.  
Such an approach can be used to generate complex visual trajectories, e.g., to avoid collisions, as done in~\cite{paolillo2022learning}.
In this work, instead, we use this formulation to enable the learned compensation terms needed to achieve the tracking of moving objects, as explained in the next section.

\section{Method}\label{sec:method}

\subsection{Problem definition}\label{sec:problem}

The aim of our work is to enable visual tracking of moving targets avoiding explicit programming of the required additional components of the basic law~\eqref{eq:vs}.

Assuming a moving target, the \ac{vs} law has to account for such motion~\cite{chaumette2007visual}:
\begin{equation}
    \bm{v} = -\lambda \widehat{\bm{L}^+}\bm{e} - \widehat{\bm{L}^+}\frac{\partial \bm{e}}{\partial t},
    \label{eq:vs_track}
\end{equation}
where the second term on the right of the equation actually acts as a feedforward term to compensate for the error's time variation due to the target motion~\cite{chaumette2007visual}.
Ad hoc techniques can be implemented to estimate the term due to the motion of the target so that it can be inserted in~\eqref{eq:vs_track} and compensated, e.g., with the introduction of integrators~\cite{chaumette1991positioning}, feedforward terms~\cite{chaumette1993tracking,corke1996dynamic} or filters~\cite{ginhoux2005active,yuen20083d}. 

In this work,  instead, our aim is to rely on an imitation strategy to infer the compensation term of the law~\eqref{eq:vs_track} from previous demonstrations of tracking experiments.
In particular, inspired by \ac{ds}-based approaches as in~\eqref{eq:ilvs}, we treat the reshaping term $\bm{\rho}$, to be learnt from data, as the compensation term in~\eqref{eq:vs_track}:
\begin{equation}
    \bm{\rho} = -\widehat{\bm{L}^+}\frac{\partial \bm{e}}{\partial t}.
    \label{eq:compensation}
\end{equation}
Therefore, our problem can be formulated as follows: learn from previous demonstrations an estimate of the compensation term $\hat{\bm{\rho}}$ so that the \ac{vs} law
\begin{equation}
    \bm{v} = -\lambda \widehat{\bm{L}^+}\bm{e} + \hat{\bm{\rho}}(\bm{e})
    \label{eq:ours}
\end{equation}
realizes tracking of moving objects. It is worth mentioning that \eqref{eq:ours} is formally the same as \eqref{eq:ilvs}. However, the vanishing term $h$ is not used in \eqref{eq:ours} since the estimate $\hat{\bm{\rho}}$ has to be always active to perform the tracking skill.  

\subsection{Dataset}\label{sec:dataset}

We assume that an ``oracle'' is available to provide a few demonstrations of the full desired tracking behavior.
A possible oracle could be a human user, who can kinesthetically teach the robot the tracking motion, or an ideal controller in simulated environments, where all the required information is perfectly known.


During the oracle's executions, data describing how the task is carried out are recorded for each timestamp.
In particular, we log the evolution of the visual error, as measured on the camera image, and the corresponding velocities, as shown to the camera in order to achieve the full desired task:
%
\begin{equation}
    {\cal D} = \left\{ \bm{e}_n^d, \bm{v}_{n}^d \right\}_{n=1,d=1}^{N,D},
    \label{eq:dataset}
\end{equation}
where $N$ is the number of samples and $D$ the number of demonstrations.
This dataset serves as the basis for the actual training set ${\cal T}$ that is built as follows:
%
%
\begin{equation}
    {\cal T} = \left\{ \bm{\varepsilon}_n^d, \bm{\rho}_n^d \right\}_{n=1,d=1}^{N,D},
    \label{eq:train_dataset}
\end{equation}
considering that $\bm{\varepsilon}_n^d = \widehat{\bm{L}^+} \bm{e}_n^d$ and $\bm{\rho}_n^d =\bm{v}_n^d +\lambda \widehat{\bm{L}^+} \bm{e}_n^d$. 
Note that for all the demonstrations we consider that the value of the control gain $\lambda$ does not change, as well as the value of the approximated inverse of the interaction matrix $\widehat{\bm{L}^+}$ is assumed to be constant and equal to its value at convergence. 

\subsection{Learning the compensation term}\label{sec:learning}


Given the training dataset~\eqref{eq:train_dataset}, 
an estimate of the compensating term can be conveniently retrieved from vision data using any regression function $\bm{r}$.
In particular, we train a \ac{gmm} on ${\cal T}$ to estimate the velocity term needed to compensate for the motion of the target object. Therefore, \ac{gmr} is used to retrieve a smooth estimate of $\bm{\rho}$, namely $\hat{\bm{\rho}}$. 
The \ac{gmr} takes as input the current value of $\bm{\varepsilon}$ and provides $\hat{\bm{\rho}}$ as
\begin{equation}
    \hat{\bm{\rho}} = \bm{r}_{\text{GMR}}(\bm{\varepsilon}~|~{\cal T }).
    \label{eq:regression}
\end{equation}
%
Therefore, the compensation term is online estimated using~\eqref{eq:regression} and inserted in the control law~\eqref{eq:ours} to achieve the tracking of moving objects.

\section{Results}\label{sec:results}

\subsection{Validation setup}\label{sec:setup}
\begin{figure}[t!]
\centering%
 \begin{subfigure}[b]{0.61\textwidth}
    \includegraphics[width=\columnwidth]{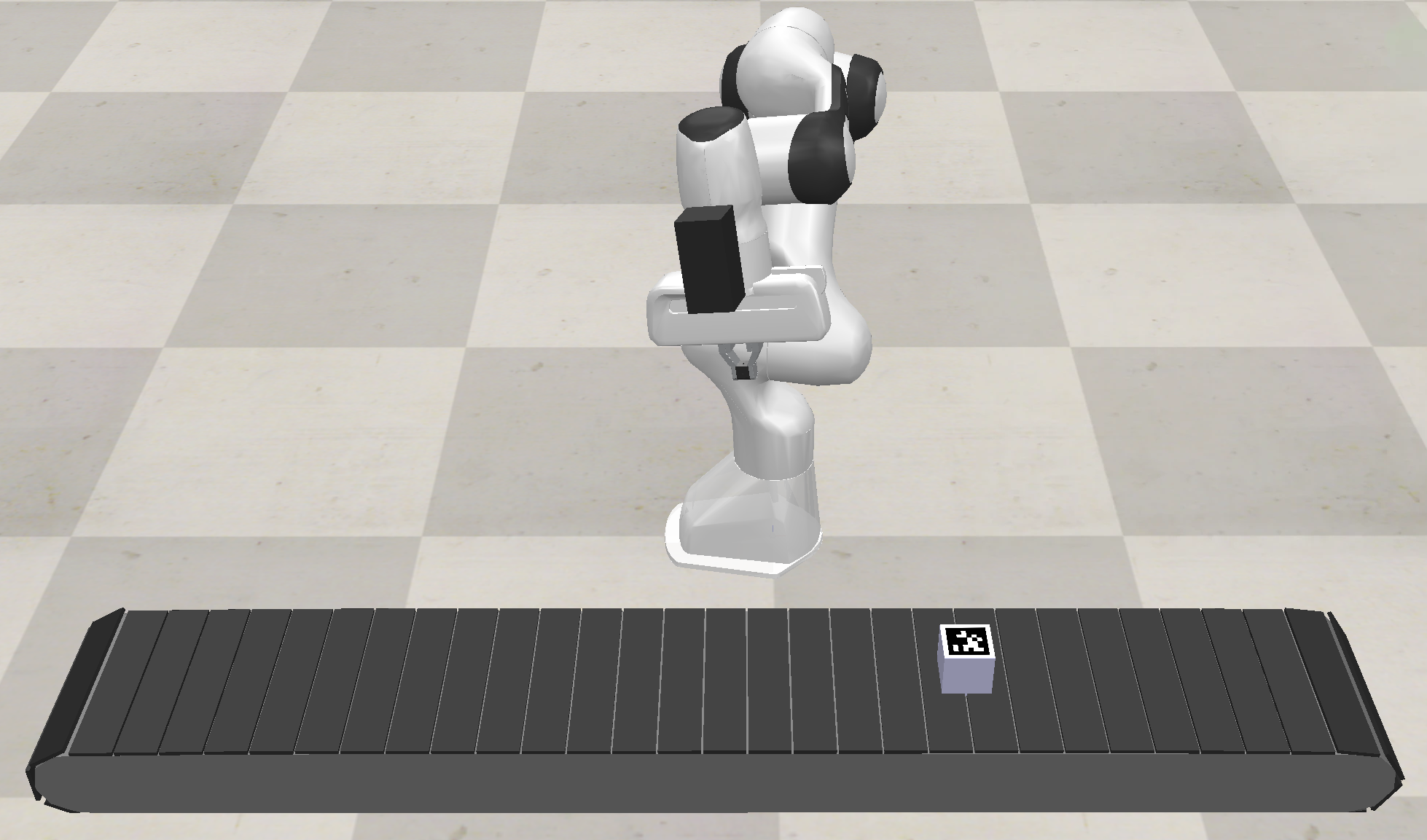}
    \caption{Front view.}
    \label{fig:front}
\end{subfigure}
\begin{subfigure}[b]{0.36\textwidth}
     \includegraphics[width=\columnwidth]{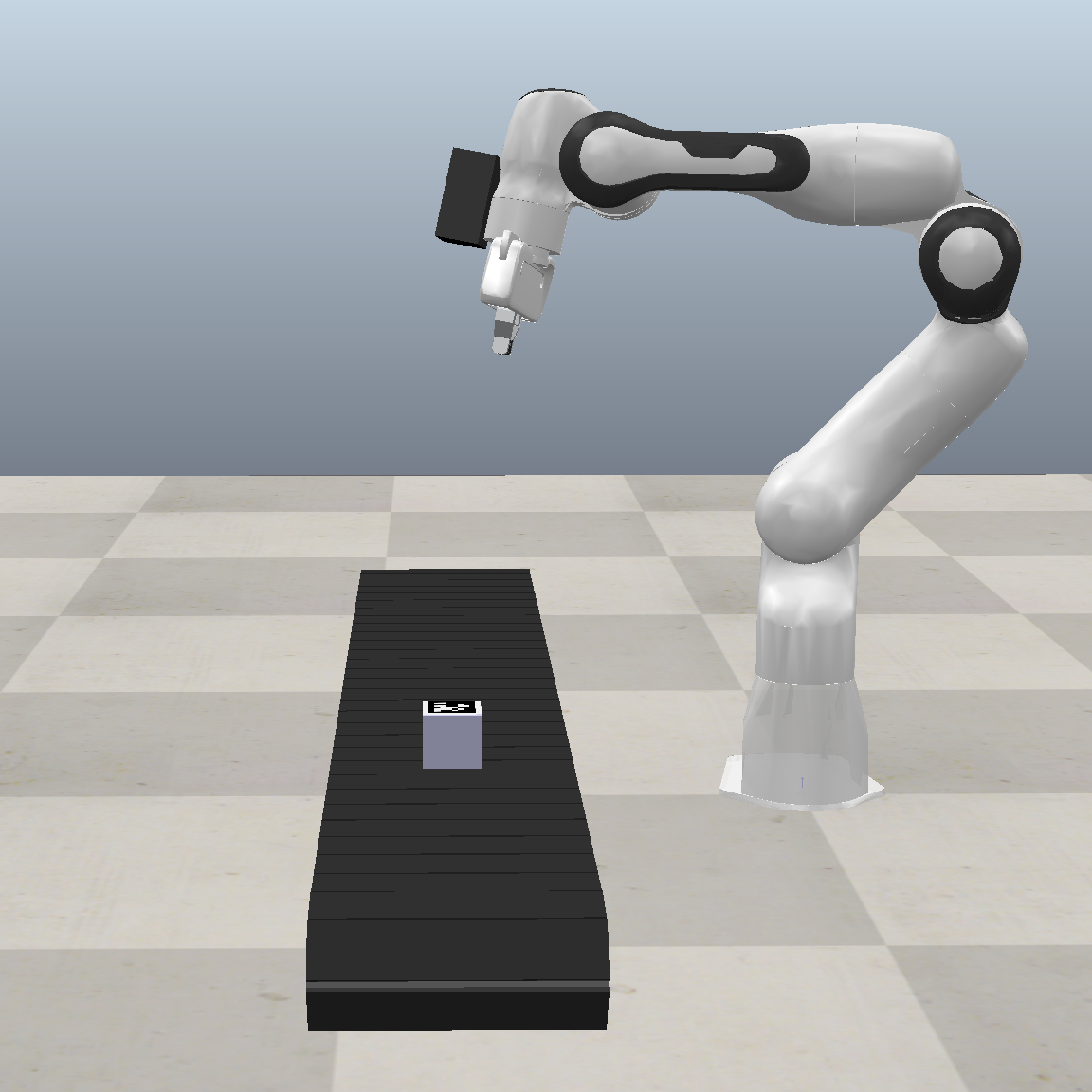}
     \caption{Side view.}
     \label{fig:side}
\end{subfigure}
\caption{Validation setup: the Franka Emika robot manipulator in the CoppeliaSim environment has to reach a box moving on a conveyor belt.}
\label{fig:setup}
\end{figure}
To validate our framework we consider a robotic experiment with the robot manipulator Franka Emika~\cite{Haddadin:ram:2022}, which has $7$~joints and an Intel RealSense D435i sensor (used as a monocular camera) mounted on the end-effector. The sensor has a field of view of 69$\degree\times$42$\degree$ and a frame resolution of 1920$\times$1080~pixel.
The robot and the environment for the experiments are simulated in CoppeliaSim~\cite{CoppeliaSim}, as shown in Fig.~\ref{fig:setup}. 
The goal of the experiment is to allow the robot to reach a box that moves at a constant velocity on a conveyor belt. 
In other terms, we set the desired features so that at convergence the robot centers the box on the image plane.
The box is marked with an AprilTag marker, whose corners provide the visual features for the \ac{vs} law. 
In particular, we use the $4$ corner points of the marker as visual features (i.e., $k=8$). As classically done in \ac{vs}, $4$ points are enough to ensure robust visual feedback.
%
At the start of the experiments, the conveyor belt accelerates from zero to $0.1$~m/s and keeps the velocity constant for the rest of the experiment. 
The implementation of the framework has been done in Python 2.7 language within the ROS~\cite{ROS} infrastructure. 

The oracle used to collect the demonstrations
consists of an ideal \ac{vs} controller provided with complete knowledge of the dynamics of the target, available in the simulated environment.
In practice, we use the law~\eqref{eq:vs_track} with $\lambda = 2$, and the compensation term is built from the perfect knowledge of the box velocity.
%
%
The interaction matrix
has been approximated by using the value of the visual features depth at the target, which is $0.09116\,$m.
In total, we have collected three demonstrations of the task.
If not otherwise mentioned, the same value of the gain and the same approximation of the interaction matrix are kept for the online experiments.
It is worth mentioning that other teaching methodologies could be used, such as kinesthetic teaching or teleoperation. 
Our choice was dictated by the need for high precision in tracking the object: a tracking controller with complete knowledge, as available in simulation, provides way better performances for precise movements than human demonstration. 
Furthermore, human demonstrations usually require preprocessing of the trajectories to grant exact convergence to the target in the feature space. 
%
The regression is carried out using \ac{gmm} with $11$ components. The number of components has been set performing a grid search.
At each iteration of the controller, the framework detects new visual features and computes the new value of $\bm{\varepsilon}$, which is used by the \ac{gmr} to compute an estimate $\hat{\bm{\rho}}$ of the compensation term that is finally inserted in the control law as in~\eqref{eq:ours}. 
%
The camera velocity thus computed is sent to the kinematic control of the manipulator that transforms it into joint velocities to move the robot towards the desired tracking behavior.
With this setup, multiple tests are carried out to evaluate firstly the learning and replication capabilities of the demonstrated target tracking tasks, and secondly, the system's ability to adapt to new scenarios and sudden changes in the environment. 

In the presented plots of the experiments, the trajectories saved in the demonstrations are shown with black dotted lines, whereas the execution of our \ac{ilvs} framework is in blue; red dots represent the starts of the demonstrated trajectories, while the red crosses are their ends.

The experiments are shown in the video accompanying the paper, available at the following link: \url{https://youtu.be/ORdAZDmCQsA}.

\subsection{Comparison with the standard \ac{vs} controller}

\begin{figure}[!t]
\centering
\setkeys{Gin}{width=\linewidth}
\begin{subfigure}{0.24\textwidth}
    \includegraphics{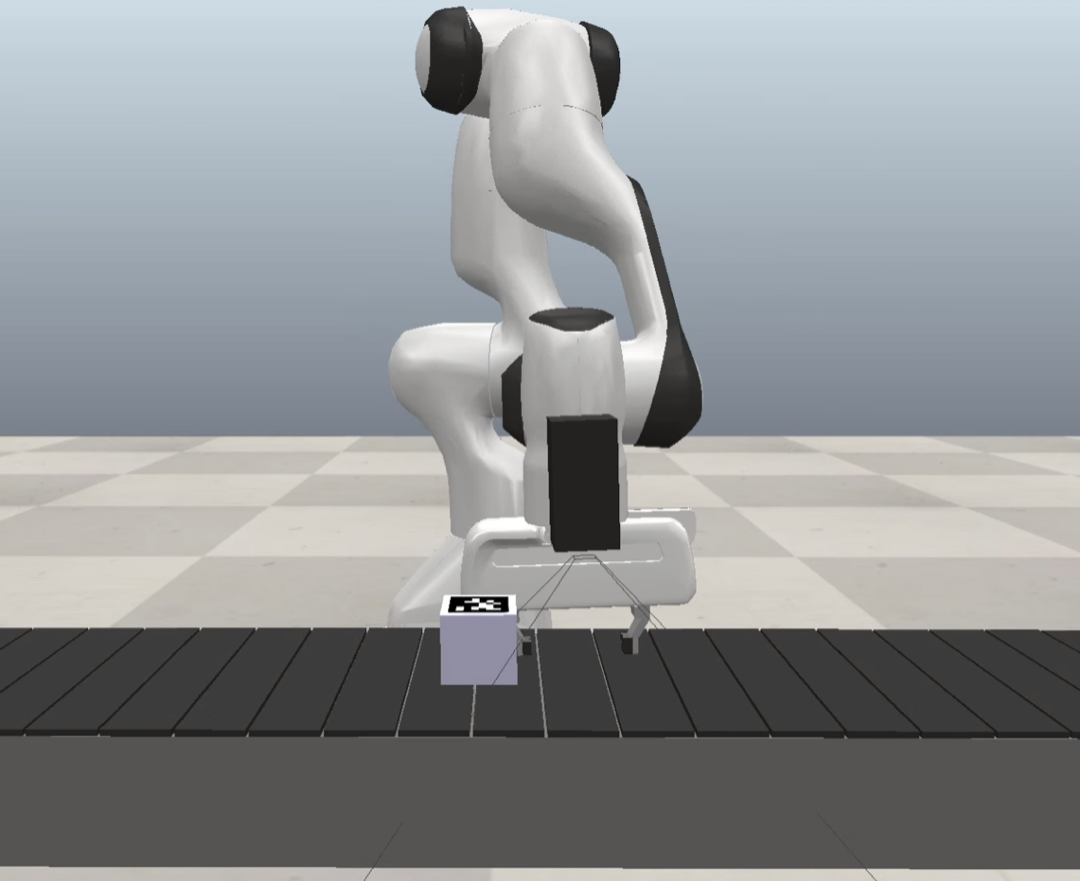}\\[3pt]
    \includegraphics{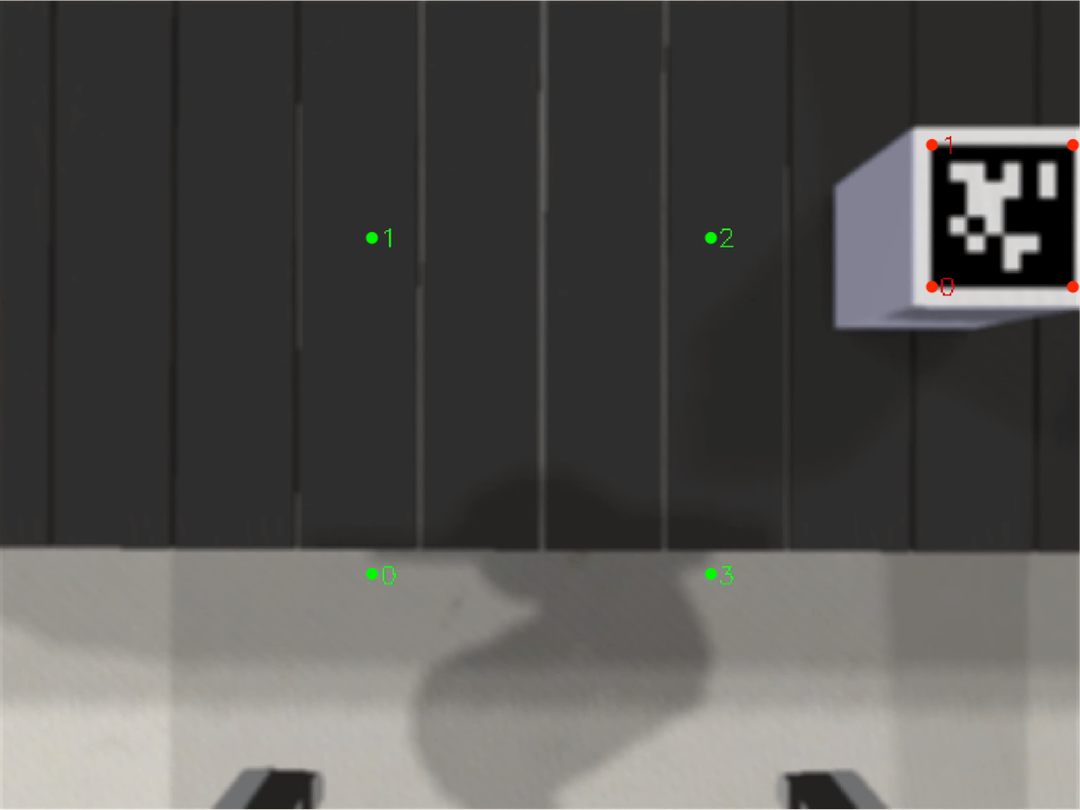}
    \caption{\ac{vs} with $\lambda = 1$}
    \label{fig:vs1}
\end{subfigure}
\hfil
\begin{subfigure}{0.24\linewidth}
    \includegraphics{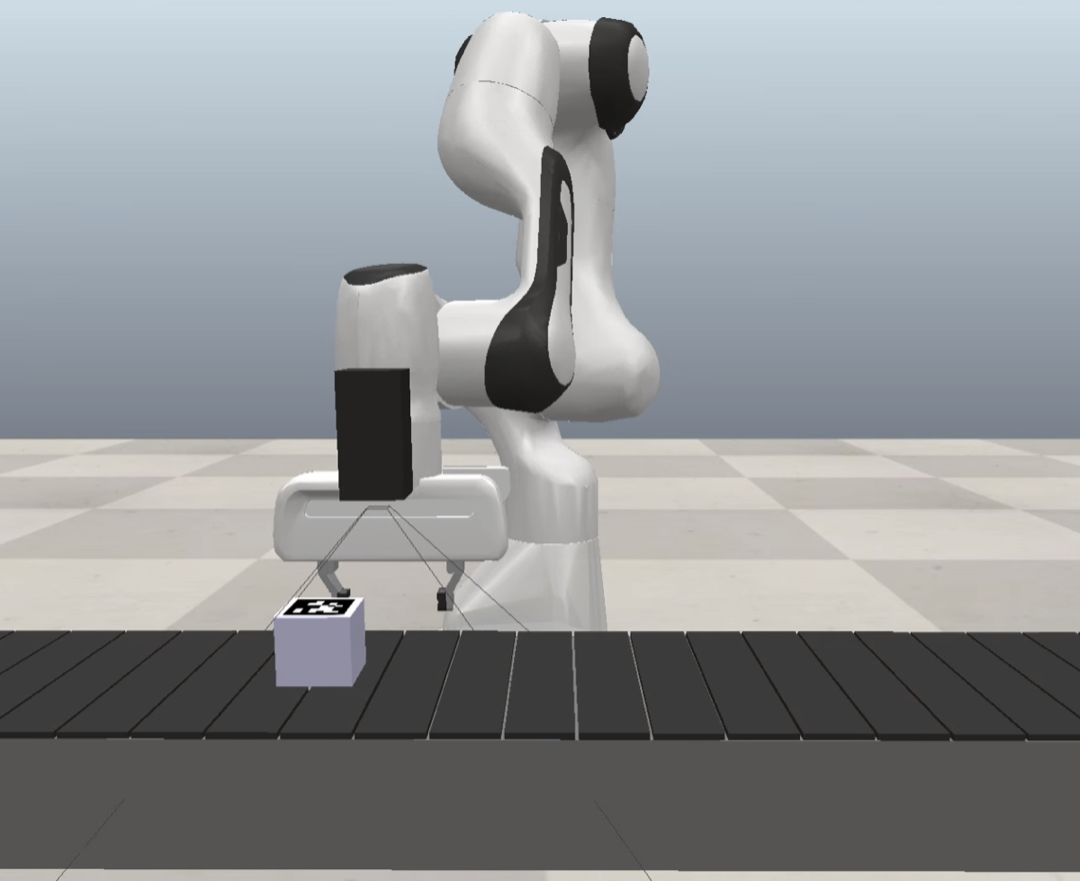}\\[3pt]
    \includegraphics{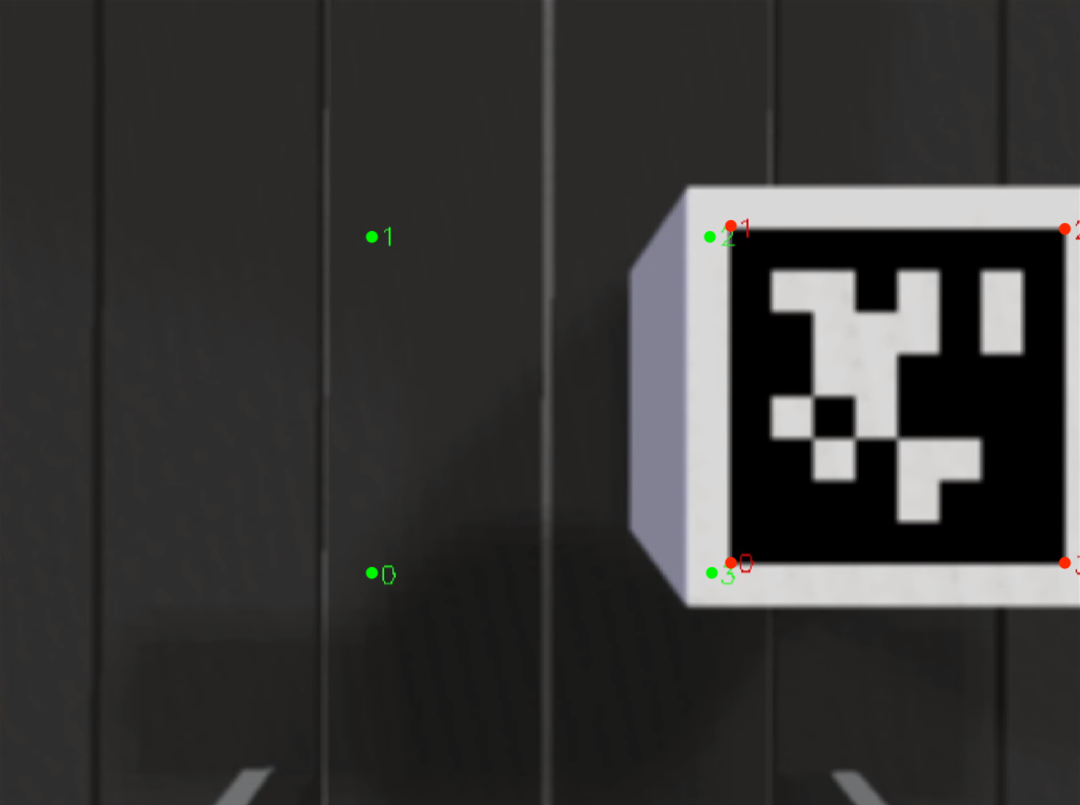}    
    \caption{\ac{vs} with $\lambda = 2$}
    \label{fig:vs2}
\end{subfigure}
    \hfil
\begin{subfigure}{0.24\linewidth}
    \includegraphics{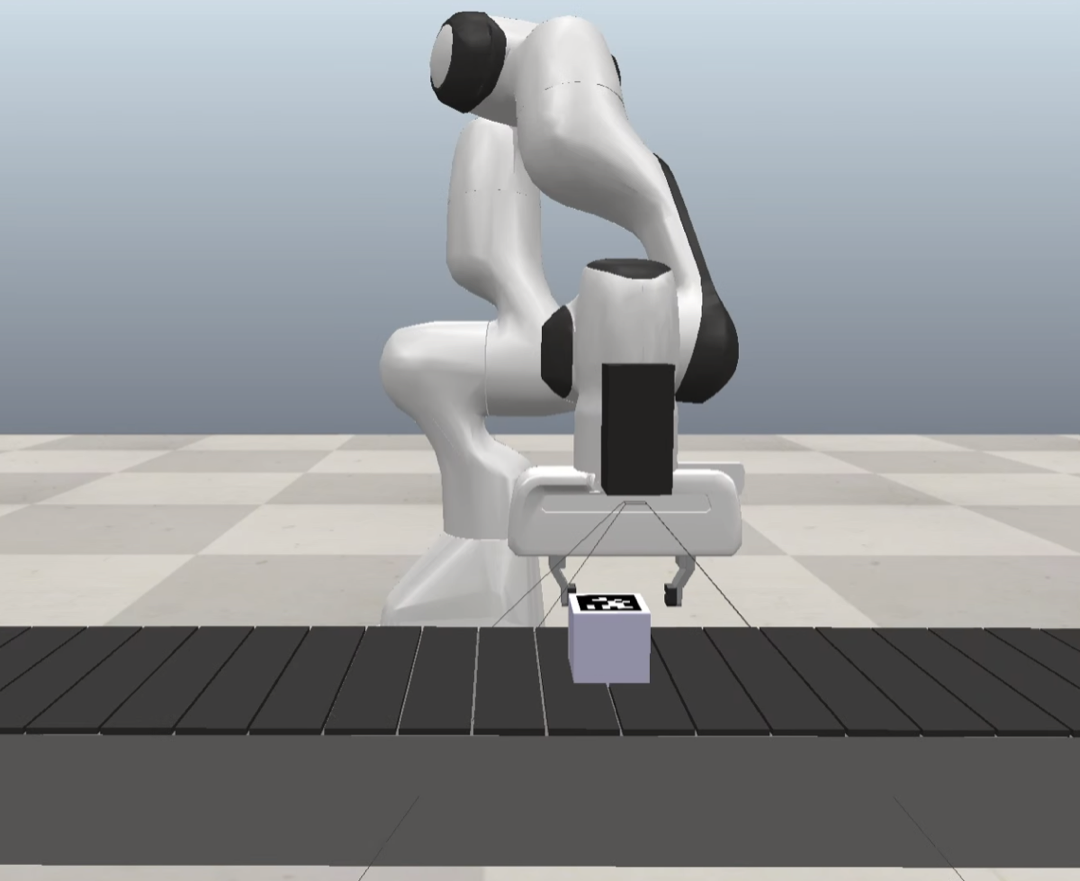}\\[3pt]
    \includegraphics{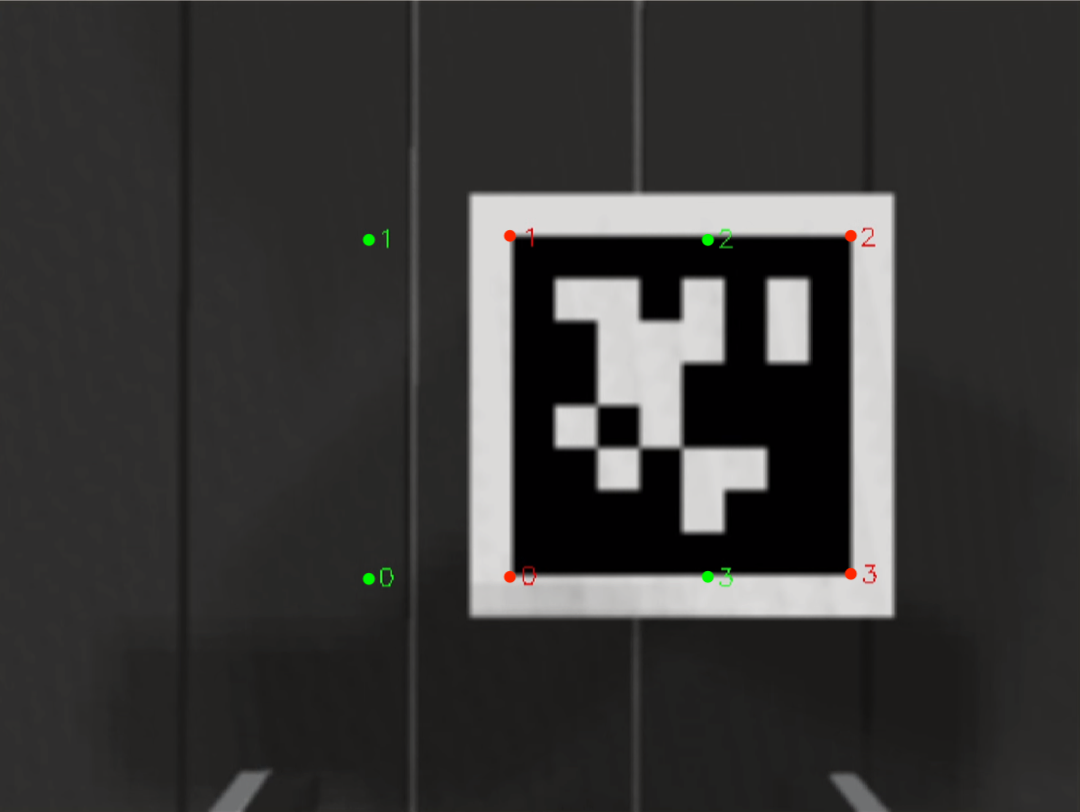}
    \caption{\ac{vs} with $\lambda = 5$}
    \label{fig:vs5}
    \end{subfigure}
\begin{subfigure}{0.24\linewidth}
    \includegraphics{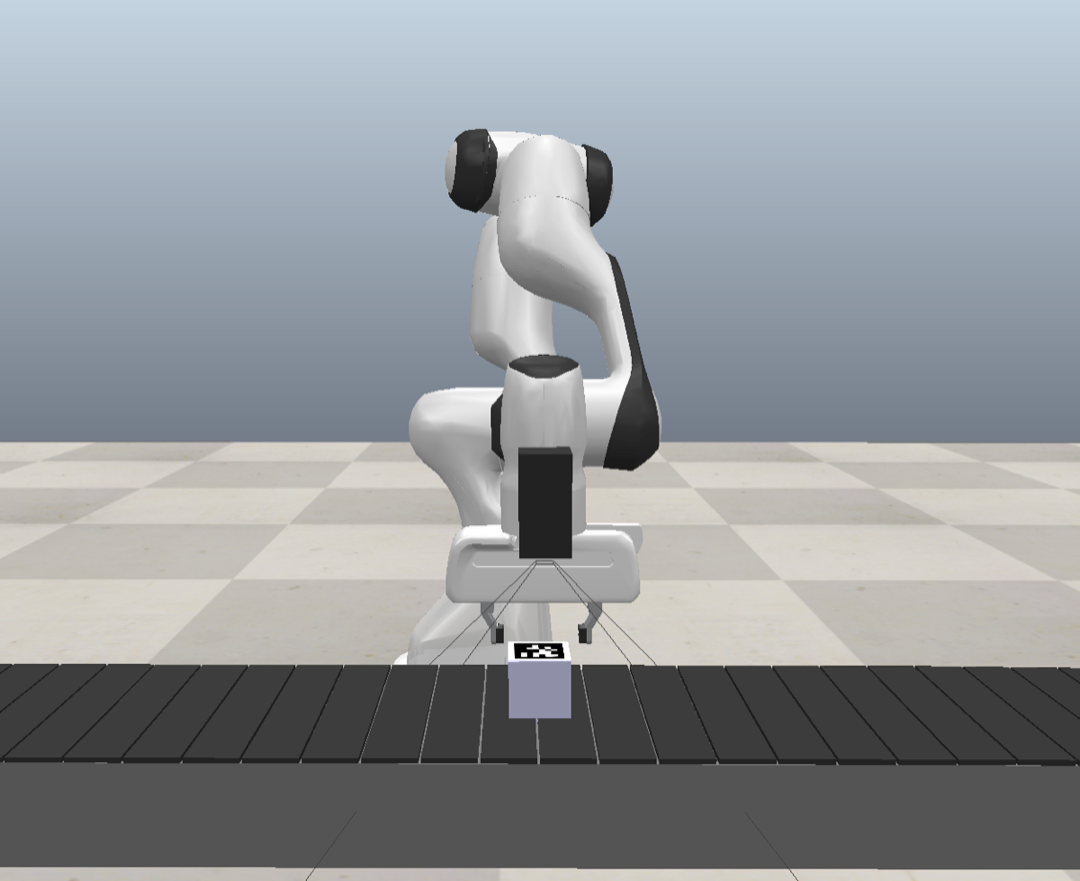}\\[3pt]
    \includegraphics{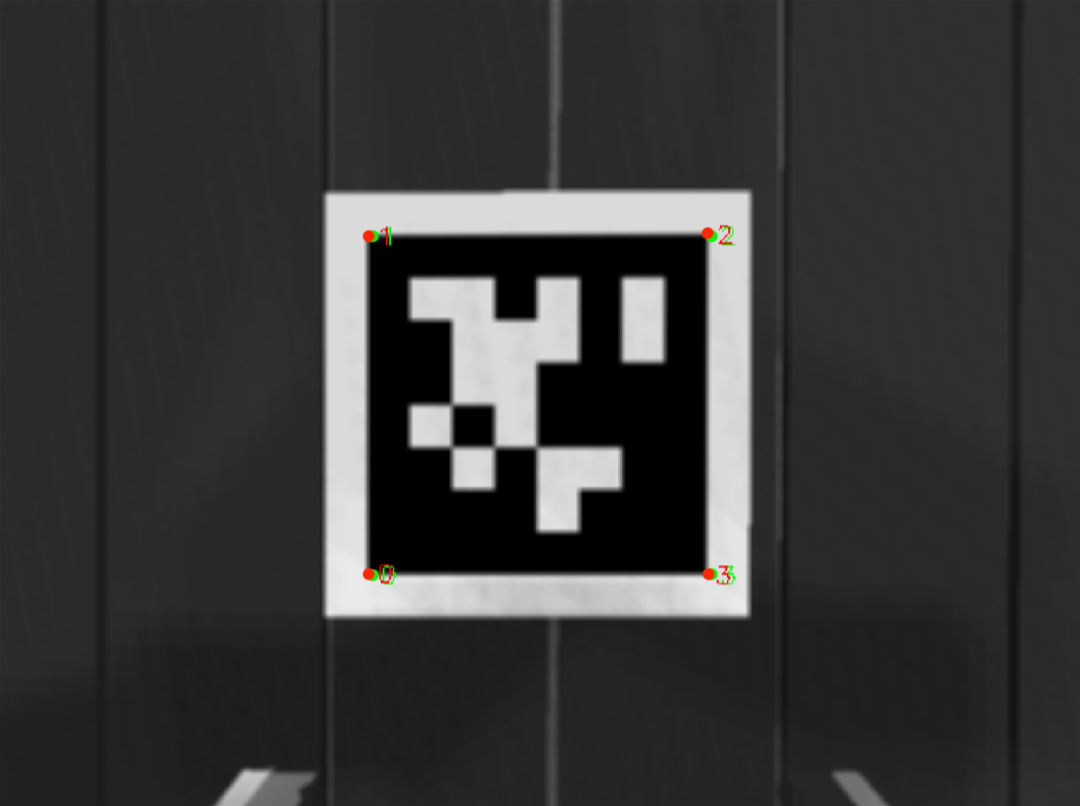}
    \caption{Proposed \ac{ilvs}}
    \label{fig:ilvs}
\end{subfigure}
    \caption{Comparison between three versions of the standard \ac{vs} controller and the proposed \ac{ilvs} strategy.}
    \label{fig:comparison}
\end{figure}
The first set of experiments aims at comparing the behavior of standard \ac{vs} without compensation term, as in~\eqref{eq:vs}, with different values of the gain $\lambda$, against our proposed \ac{ilvs} strategy.
The results of this comparison are shown in Fig.~\ref{fig:comparison}.
%
As expected, even if the standard \ac{vs} law manages to approach the box, due to its motion, it never manages to center it on the image plane. 
Indeed, a constant error between the current state of the features (denoted in red and numbered from zero to three in Fig.~\ref{fig:comparison}) and their desired position (in green) is kept at a steady state. 
Such error is lower by increasing the value of $\lambda$ from $1$ to $5$, but cannot be nullified. 
It is indeed noteworthy that extremely high gain values cannot provide a reasonable solution to the tracking problem, since it would introduce instability in the control system~\cite{chaumette2006visual,chaumette2007visual}. 
%
Unlike the standard controllers, our \ac{ilvs} manages to infer from data the required information to compensate for the box motion.
As shown in Fig.~\ref{fig:ilvs}, \ac{ilvs} provides the robot with the capability to approach the target, reach convergence, and keep the camera above the box at the desired pose for the duration of the experiment.
Indeed, in this case, the measured visual features match their desired counterpart at steady-state. 

\begin{figure}[!t]
  \includegraphics[width=0.65\columnwidth]{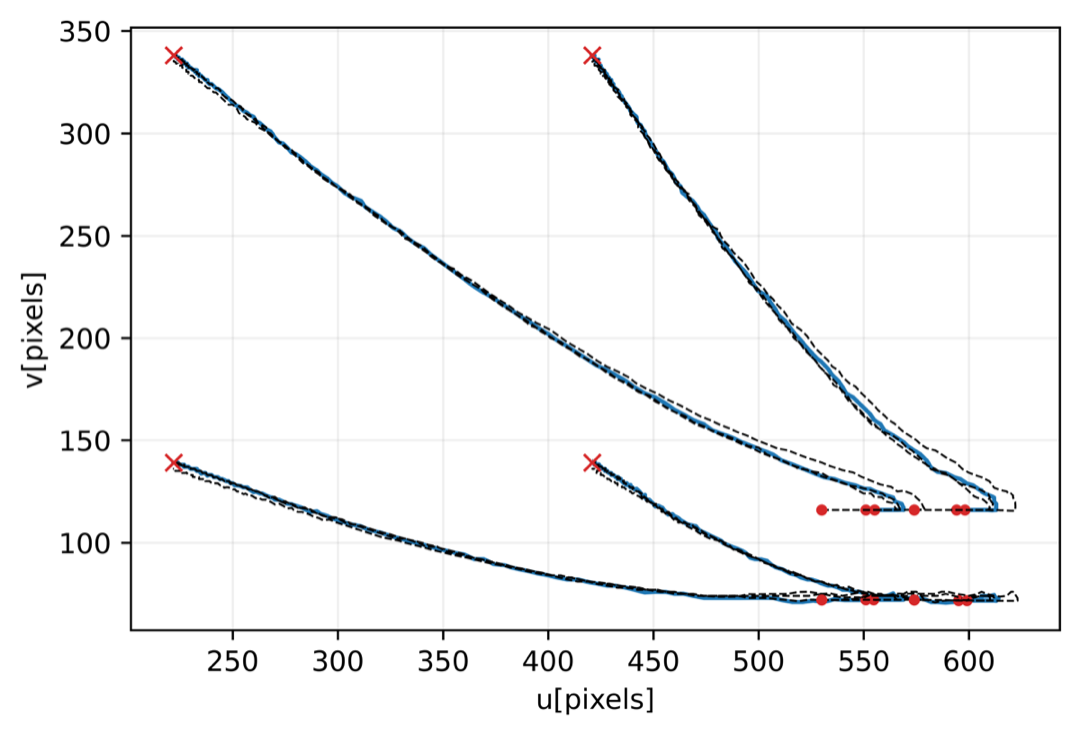}%
  \caption{\ac{ilvs} experiment with the same initial condition as in the demonstration: visual features trajectories as in the demonstrations and executed by our method.}
  \label{fig:same_pos}
\end{figure}
Fig.~\ref{fig:same_pos} shows, for the same experiment, a qualitative evaluation of the trajectories of the visual features from the demonstrations (black dotted lines), and the trajectories executed by the \ac{ilvs} strategy (in blue).
One can observe the ability of the system to accurately replicate the demonstrated trajectories when starting from a known location (the same as the demonstrated ones).
%
%
    
%
The correspondent quantitative results of this experiment are presented in terms of average \ac{rmse}\footnote{\ac{rmse} values rounded up to the nearest whole number.} and its standard deviation measuring the accuracy of the predicted camera position and velocity, and the predicted feature position w.r.t the corresponding quantities contained in the demonstrations.
In particular, the average \ac{rmse} regarding the predicted visual features position is $22 \pm 11$~pixel. For the camera positions and the linear camera velocities, the obtained results are $33 \pm 24$~mm and $69 \pm 71$~mm/s, respectively.



\subsection{Target tracking experiments with unseen initial conditions}

The second set of experiments is carried out to test the adaptability of the system w.r.t. unseen initial conditions, i.e.,
when 
the starting orientation or the position of the camera is different from those demonstrated in the training dataset. 

\begin{figure}[t!]
\centering%
 \begin{subfigure}[b]{0.49\textwidth}
 \centering
    \includegraphics[width=\columnwidth]{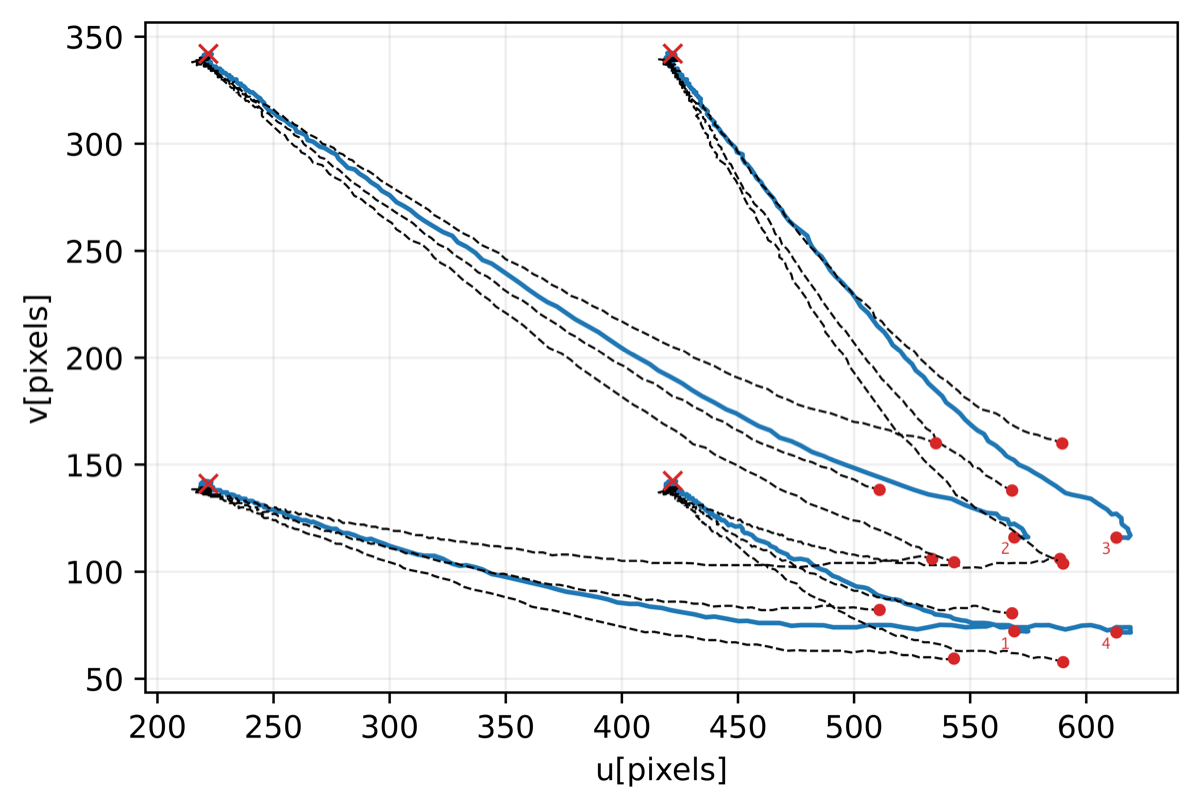} 
\end{subfigure}
\hfill
\begin{subfigure}[b]{0.5\textwidth}
     \includegraphics[width=\columnwidth]{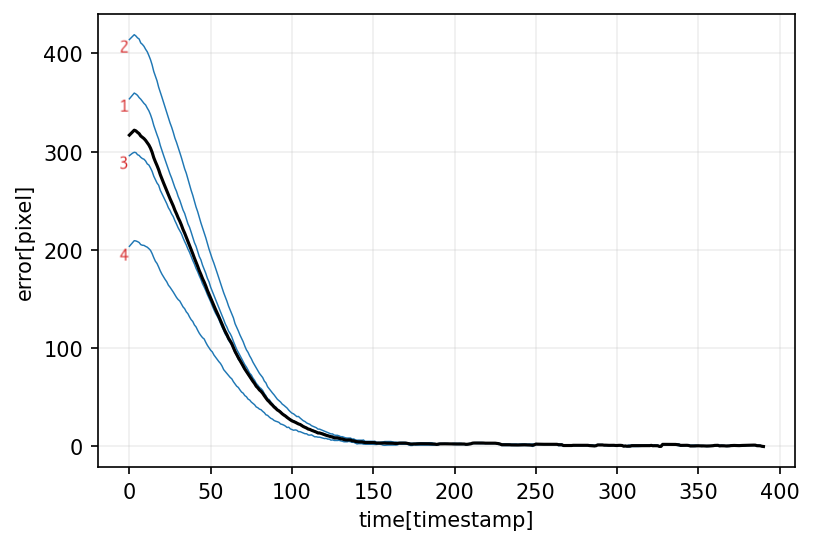}%
\end{subfigure}
\caption{\ac{ilvs} experiment with similar initial conditions of the demonstrated ones: visual features trajectories~(left) and visual error~(right).}
\label{fig:exp1}
\end{figure}

\begin{figure}[btp]
\centering
\setkeys{Gin}{width=\linewidth}
\begin{subfigure}{0.24\textwidth}
    \includegraphics{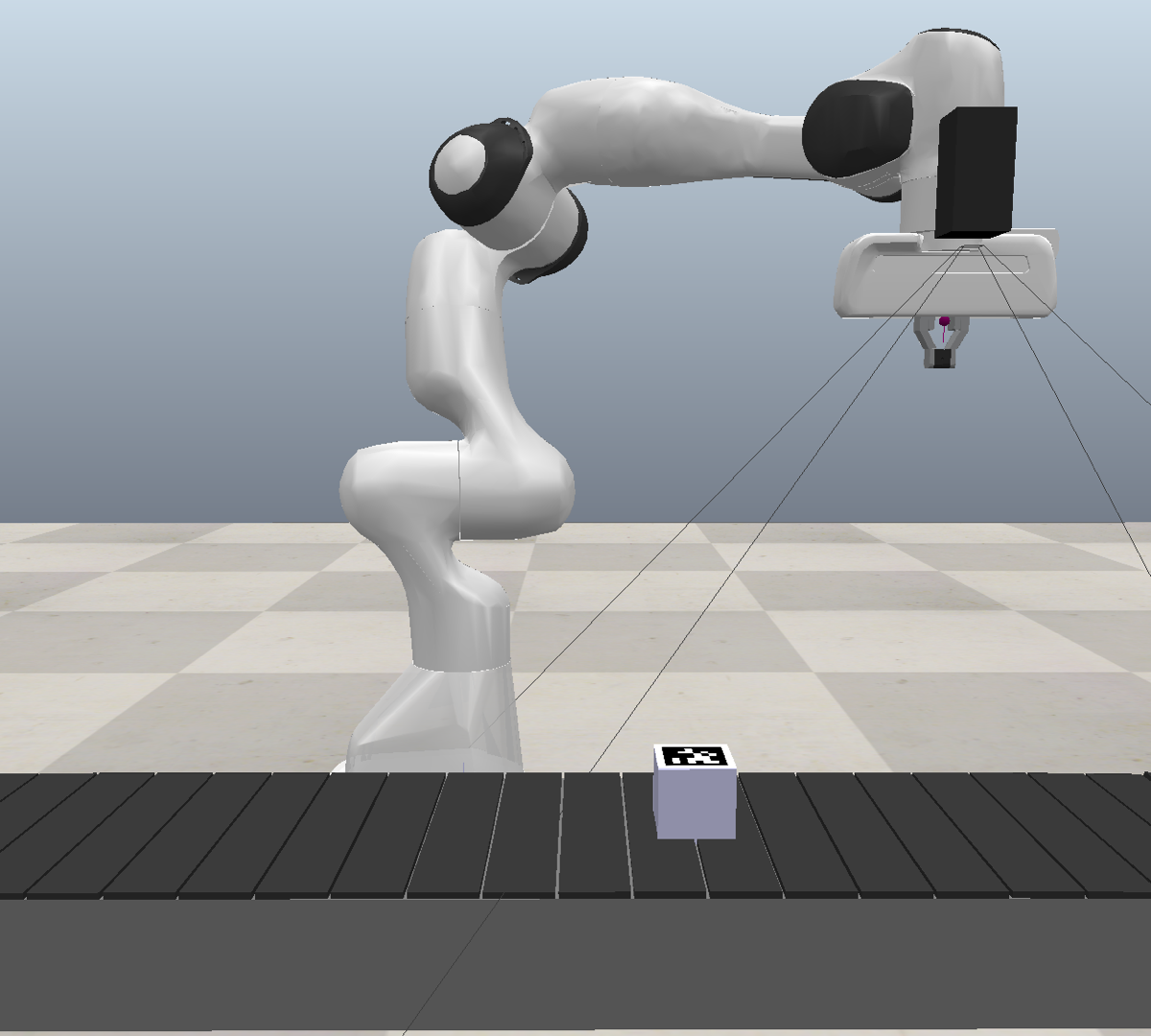}\\[3pt]
    \includegraphics{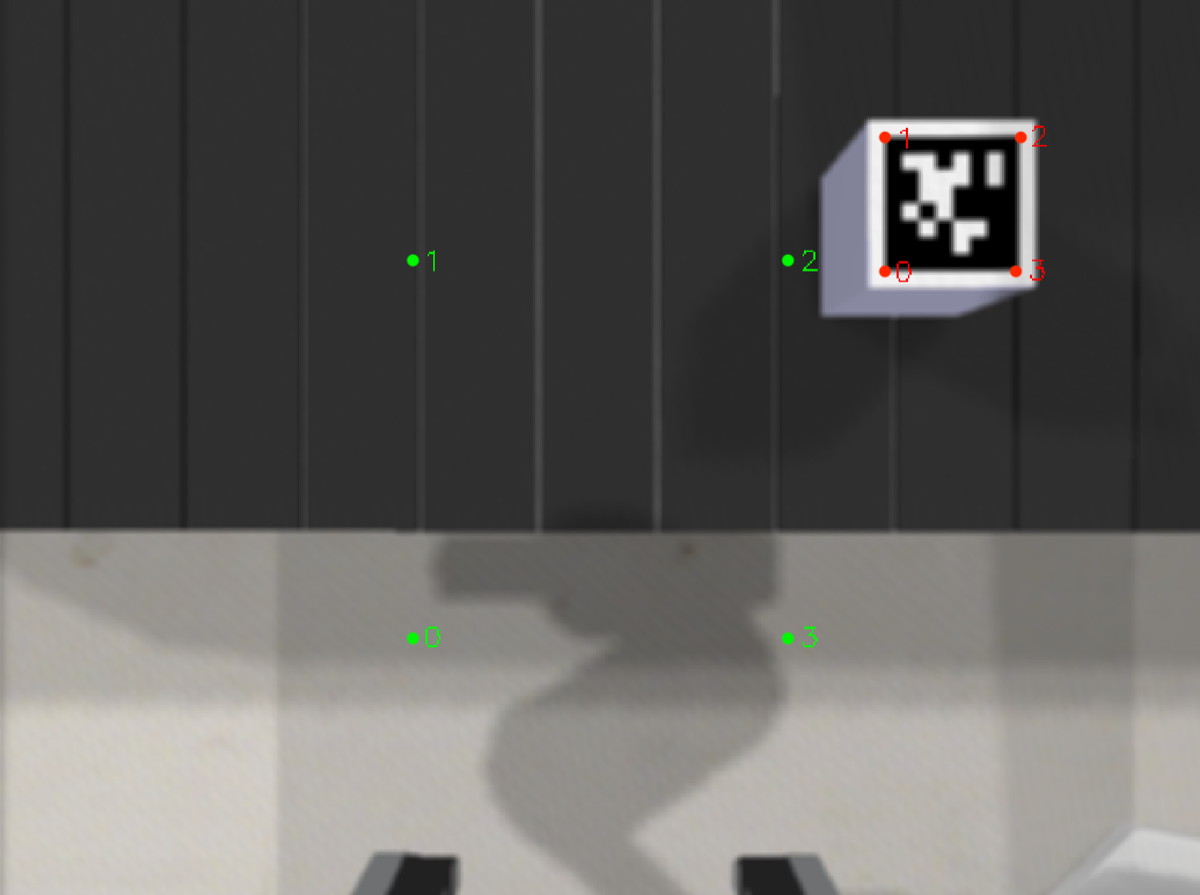}
\end{subfigure}
\hfil
\begin{subfigure}{0.24\linewidth}
    \includegraphics{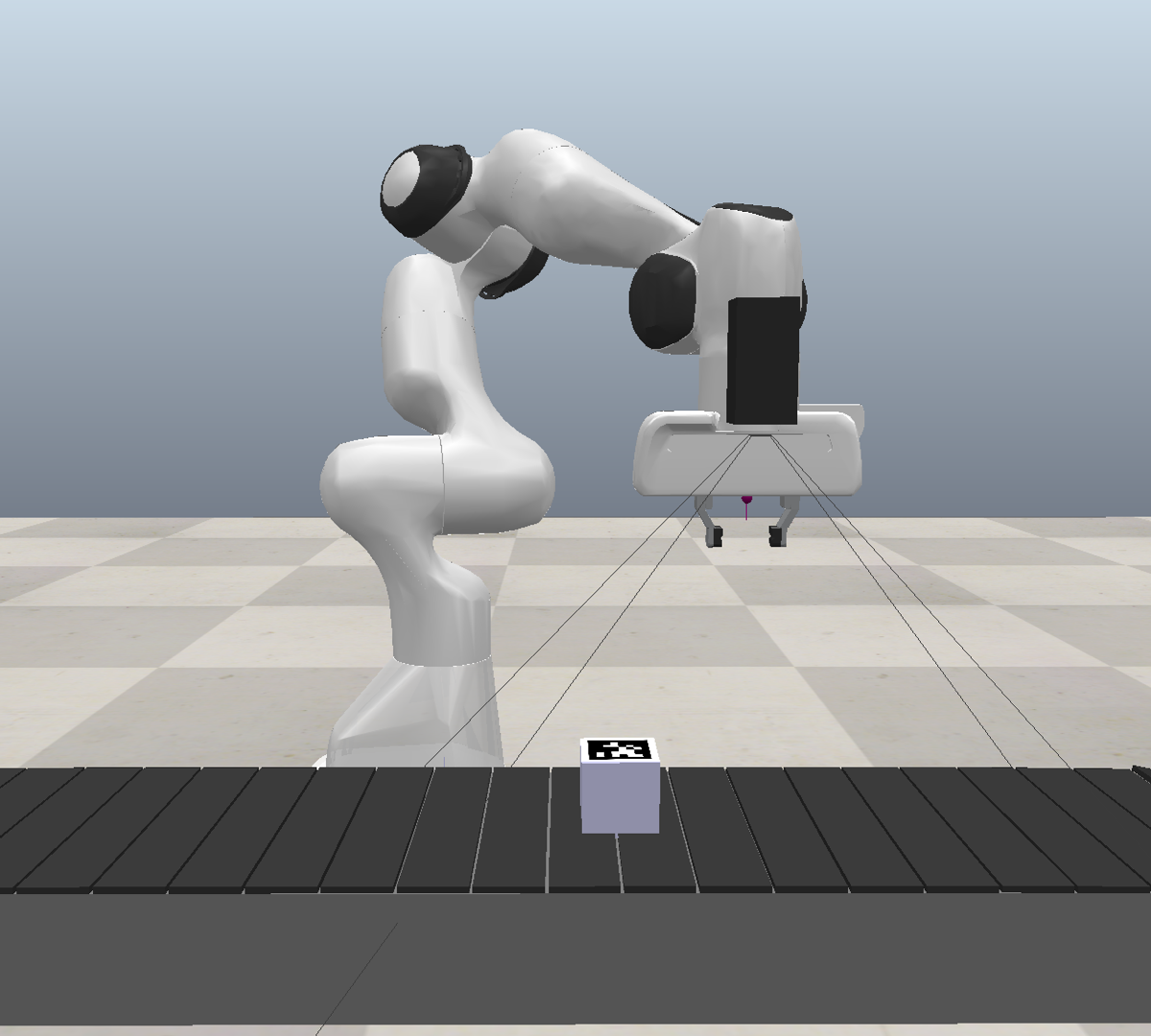}\\[3pt]
    \includegraphics{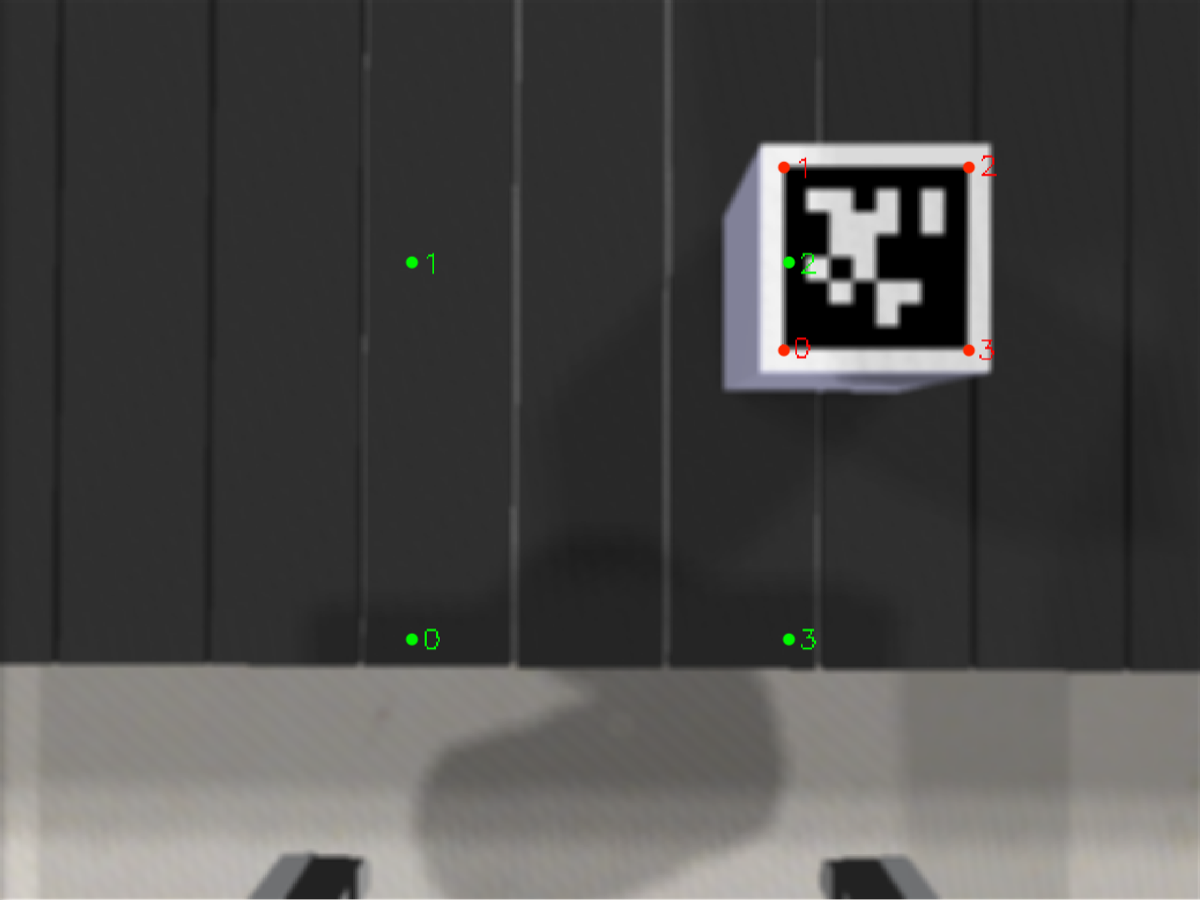} 
\end{subfigure}
\hfil
\begin{subfigure}{0.24\linewidth}
    \includegraphics{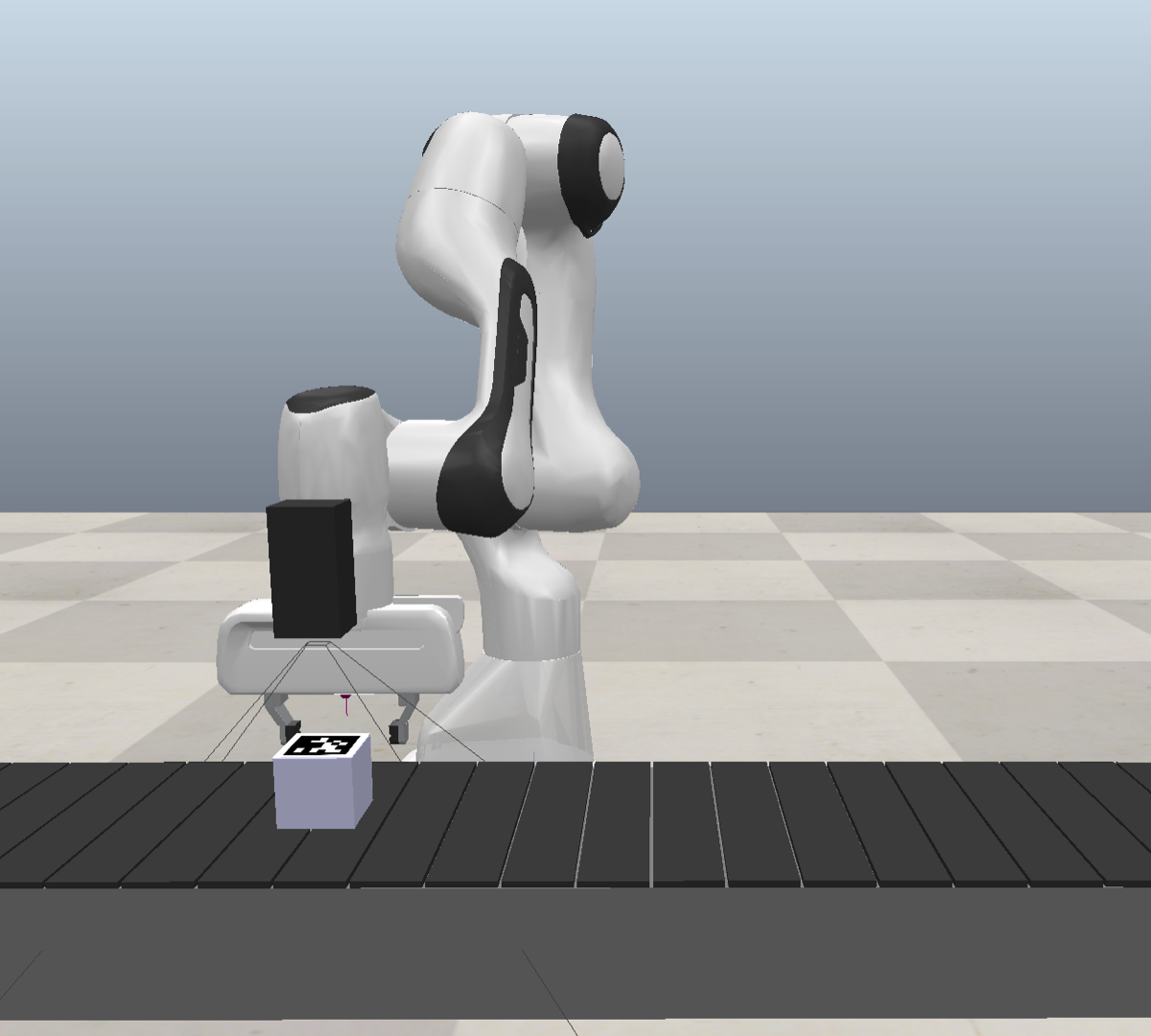}\\[3pt]
    \includegraphics{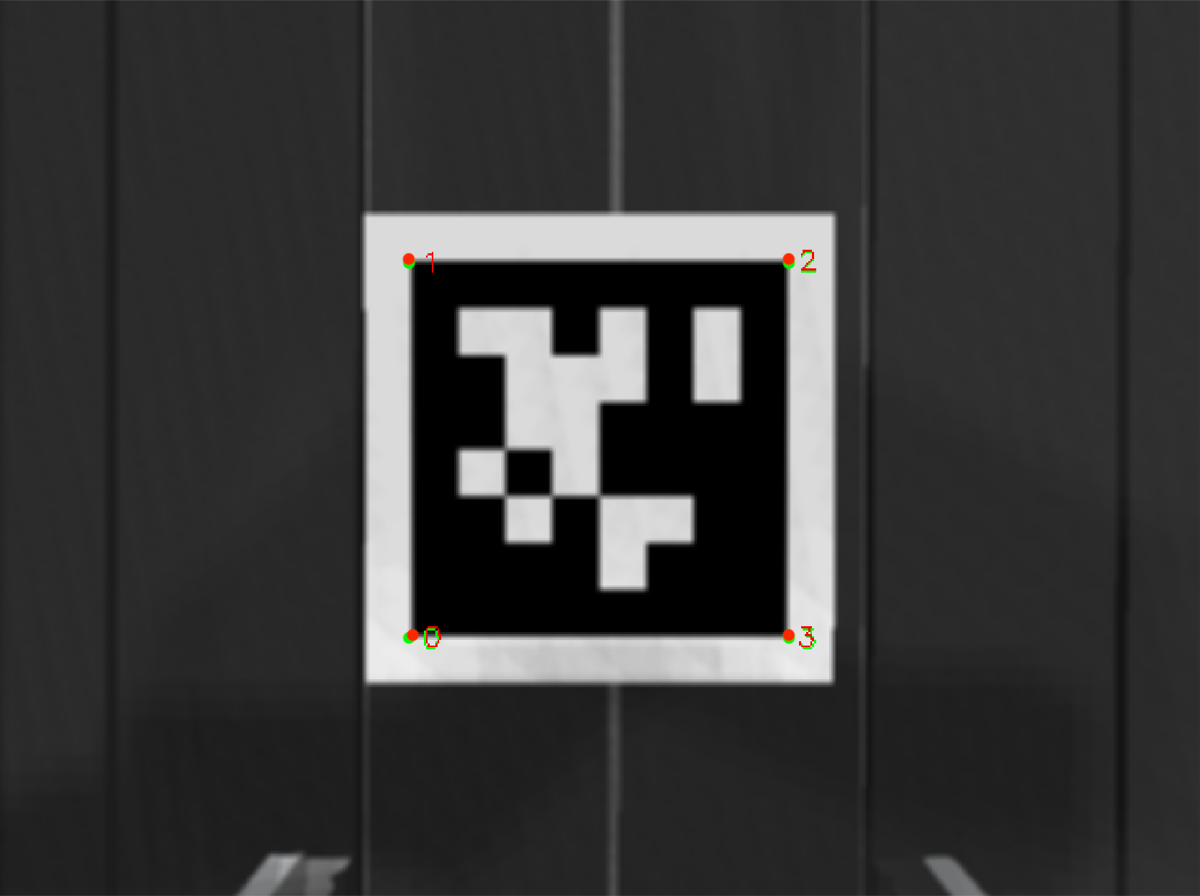}
\end{subfigure}
\hfil
\begin{subfigure}{0.24\linewidth}
    \includegraphics{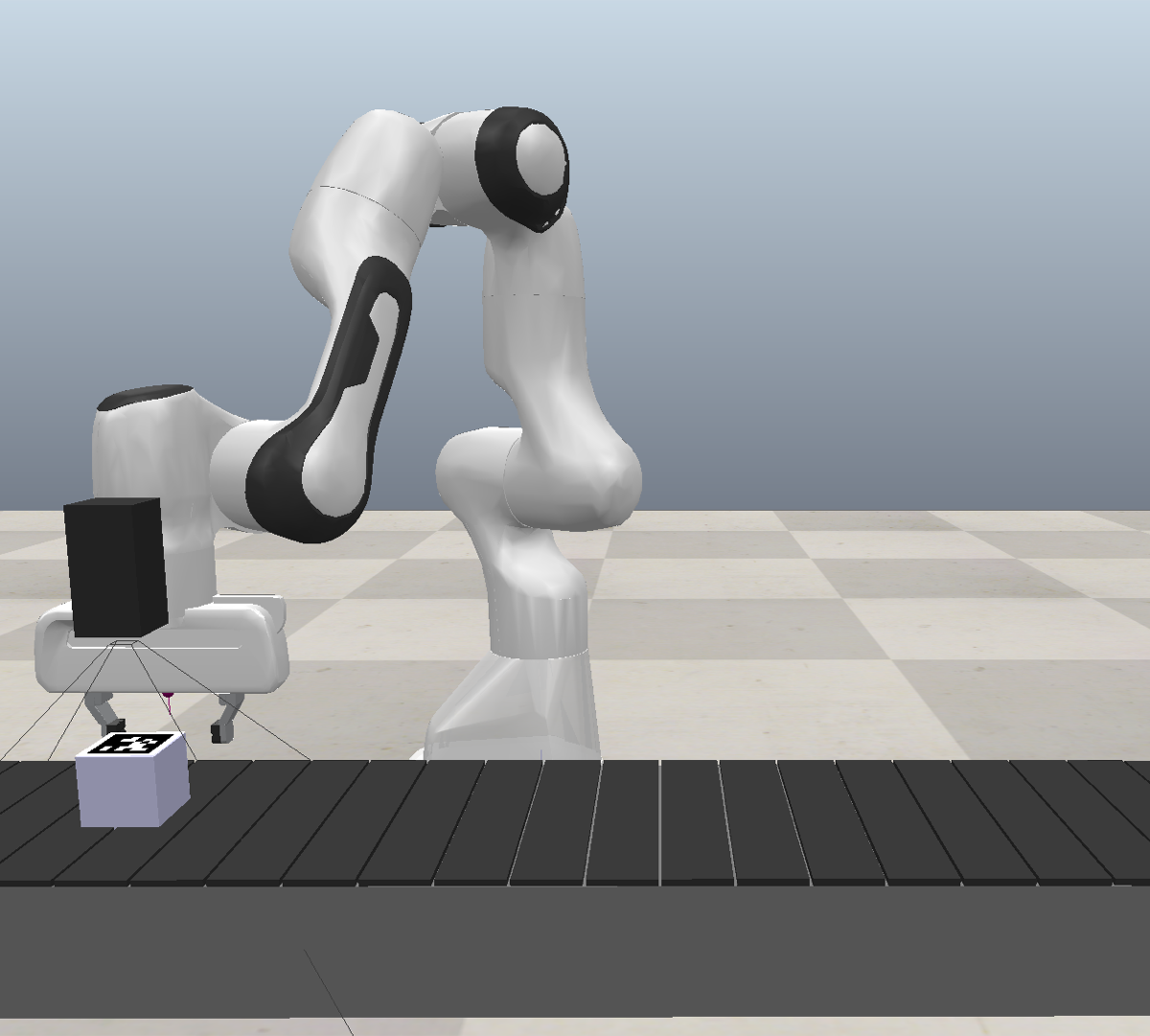}\\[3pt]
    \includegraphics{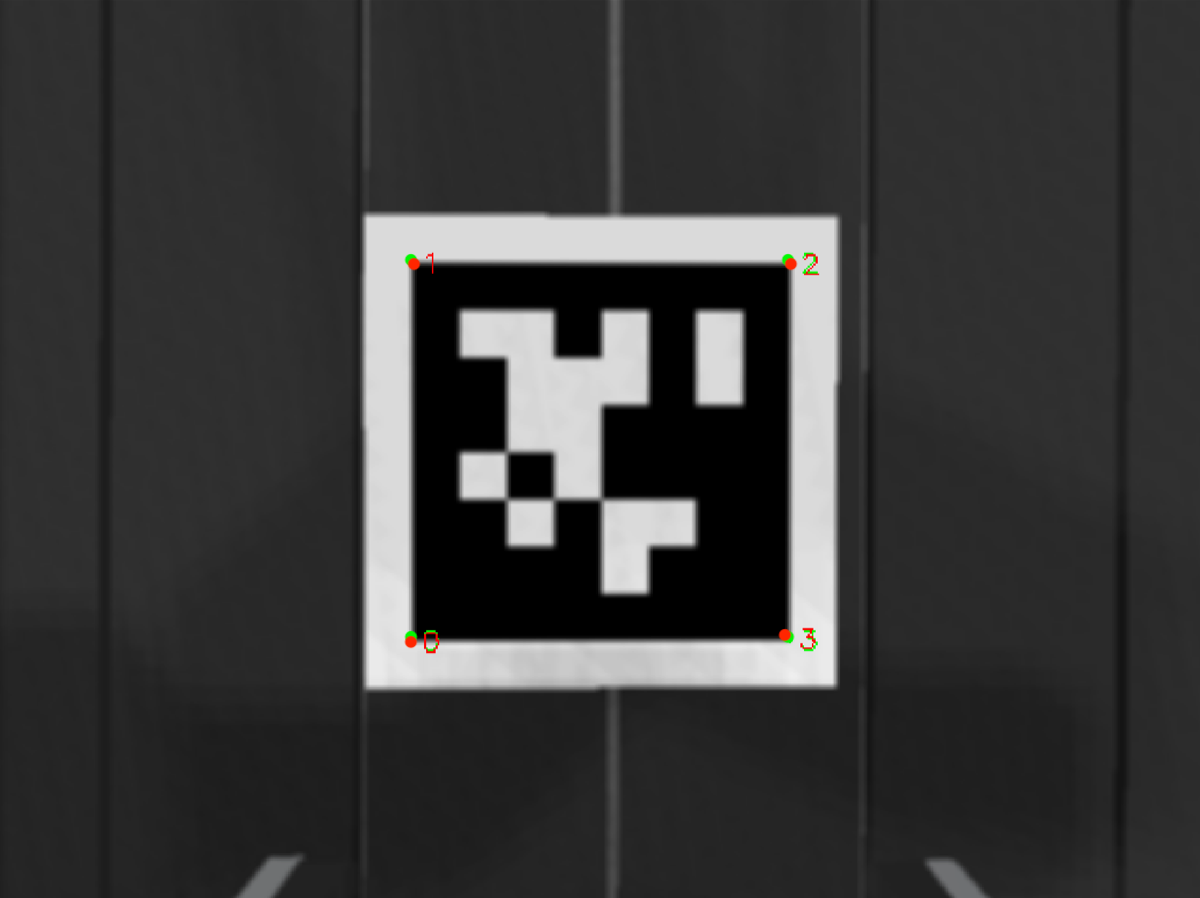}
\end{subfigure}
 \caption{Snapshots of the \ac{ilvs} experiment with similar initial conditions of the demonstrations: robot's external views (top) and camera images (bottom).} 
    \label{fig:adapt1}
\end{figure}

We tested the framework with incremental levels of difficulty.
In the first experiment of this set, the initial conditions are analogous (but not identical as in the experiment shown in Fig.~\ref{fig:ilvs} and Fig.~\ref{fig:same_pos}) to the ones in the training dataset.
As illustrated in Fig.~\ref{fig:exp1} (left), the starting point of the experiment in the image plane are in the nearby of the starting points (red dots) of the demonstrations (black dotted lines), since the initial position of the camera has been slightly moved away from the one in the demonstrations.
The starting orientation of the camera is, instead, the same as the demonstrations.
Given similar initial conditions, as expected, the system executes the task (blue lines in the plot) without any particular difficulties.
Fig.~\ref{fig:exp1} (right) shows the time evolution of the visual error for each of the four features (blue lines), which is kept to zero after a transient time for the duration of the experiment; it is also depicted the average visual error among all features (black line).
Four snapshots of this experiment are presented in Fig.~\ref{fig:adapt1} showing the manipulator approaching the object and tracking the target moving on the conveyor belt during all its motion. 
%
%

\begin{figure}[t!]
\centering%
 \begin{subfigure}[b]{0.5\textwidth}
    \includegraphics[width=\columnwidth]{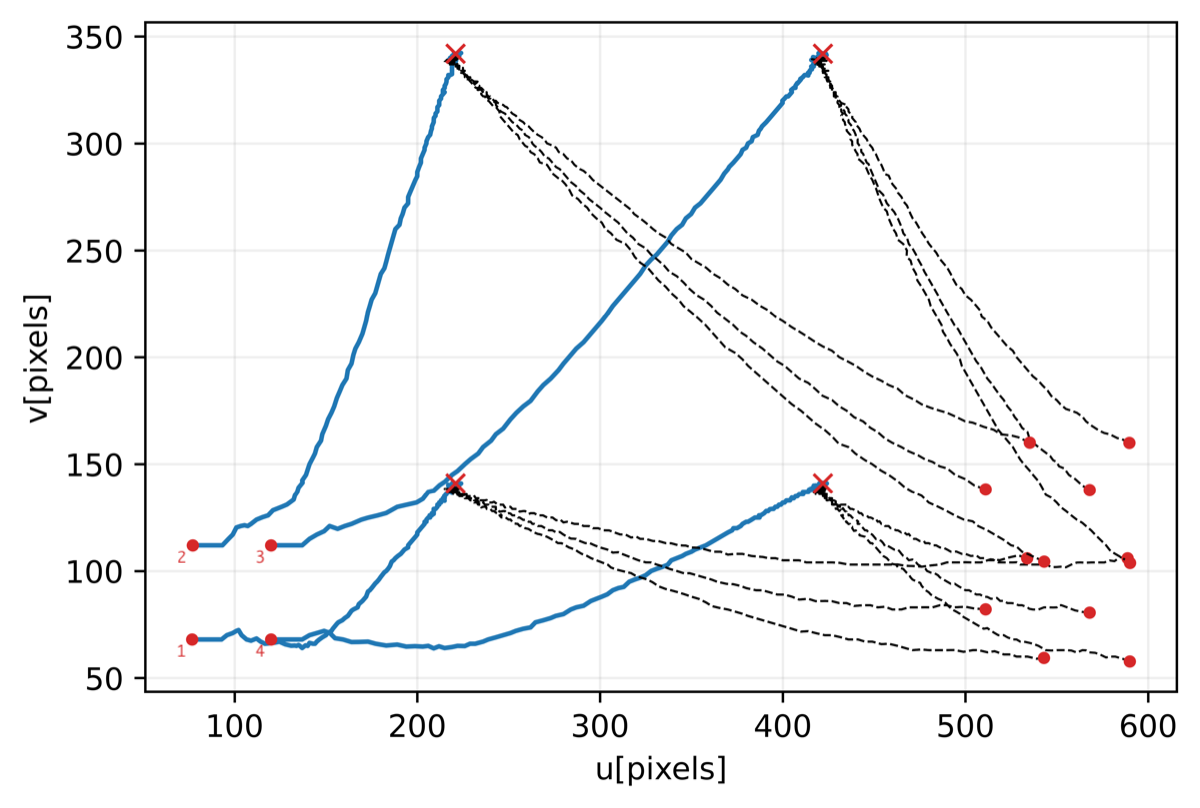} 
\end{subfigure}%
\hfill
\begin{subfigure}[b]{0.5\textwidth}
     \includegraphics[width=\columnwidth]{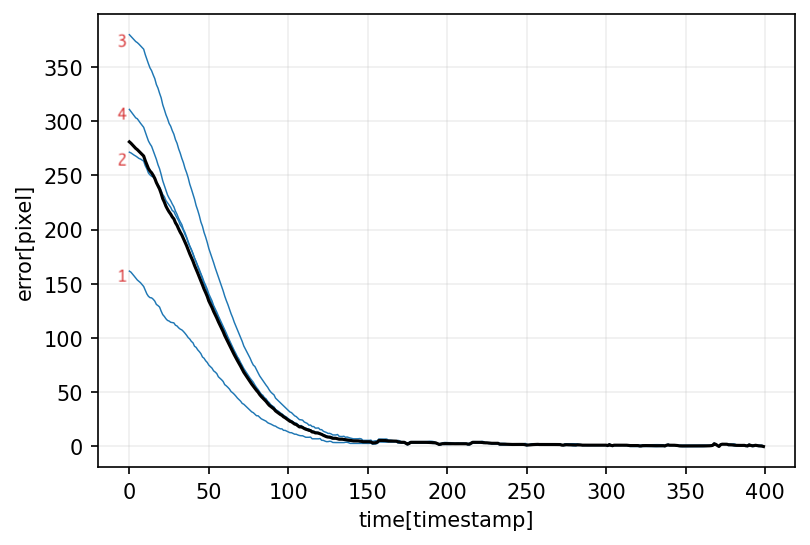}%
\end{subfigure}%
\caption{\ac{ilvs} experiment with unseen initial position and initial orientation as in the demonstrations: visual features trajectories~(left) and visual error~(right).}%
\label{fig:exp2}%
\end{figure}

\begin{figure}[!t]%
\centering%
\setkeys{Gin}{width=\linewidth}%
\begin{subfigure}{0.24\textwidth}%
    \includegraphics{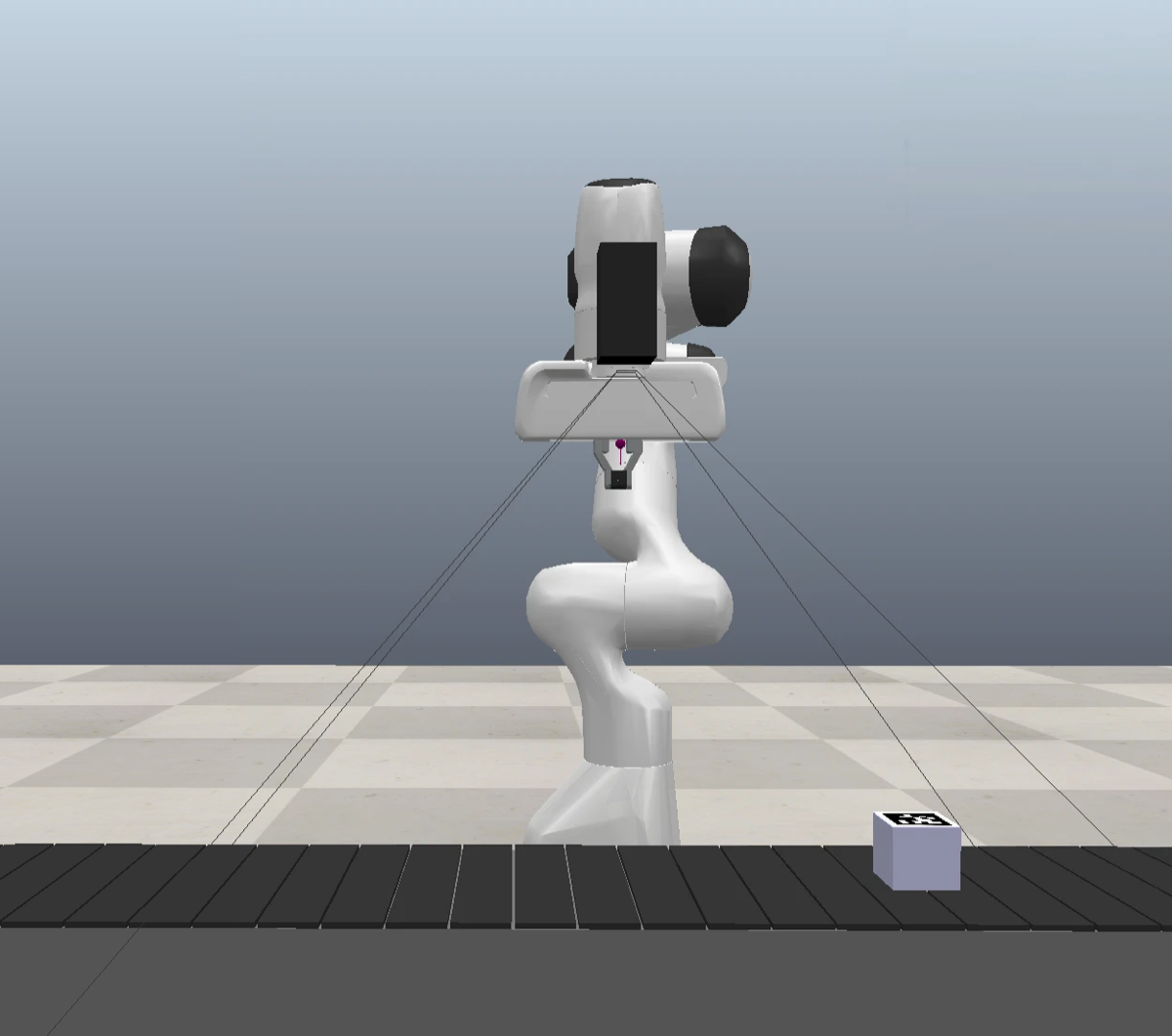}\\[3pt]
    \includegraphics{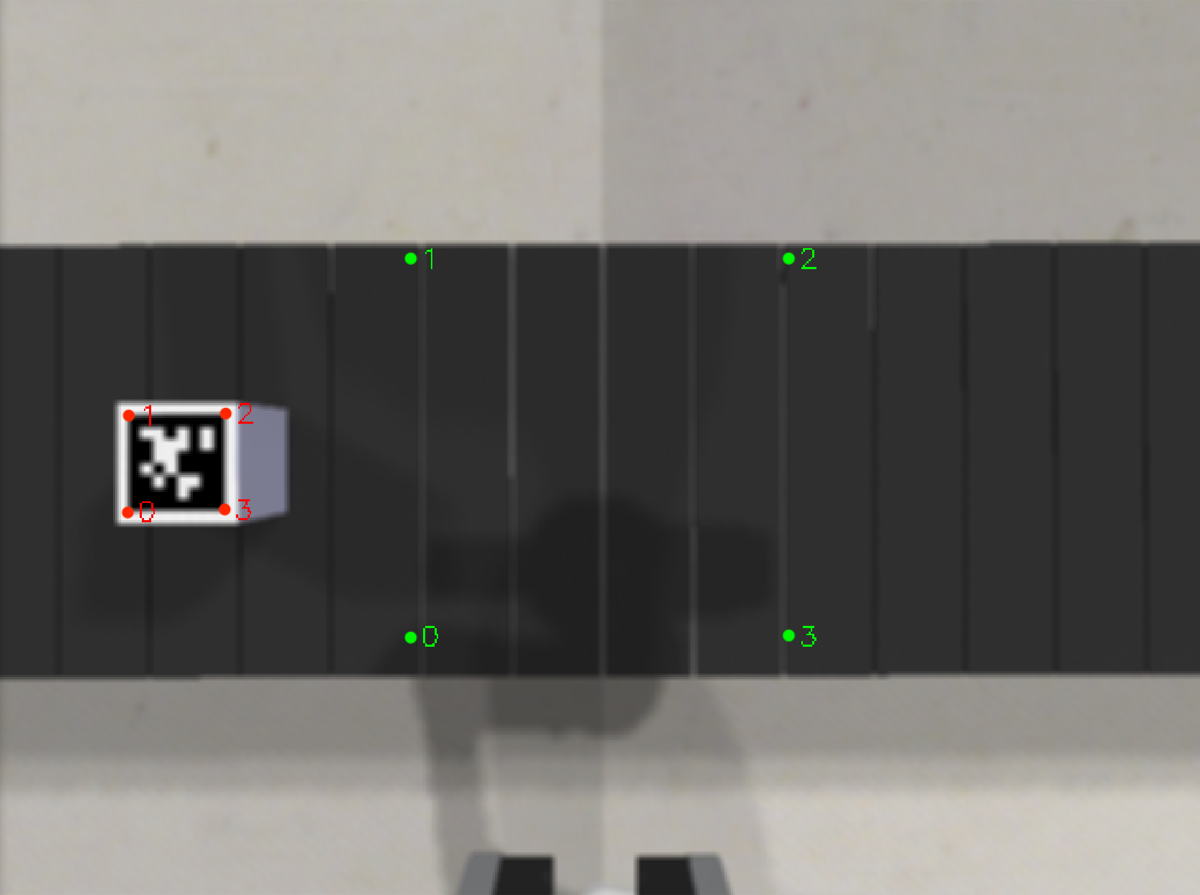}
\end{subfigure}
\hfil
\begin{subfigure}{0.24\linewidth}
    \includegraphics{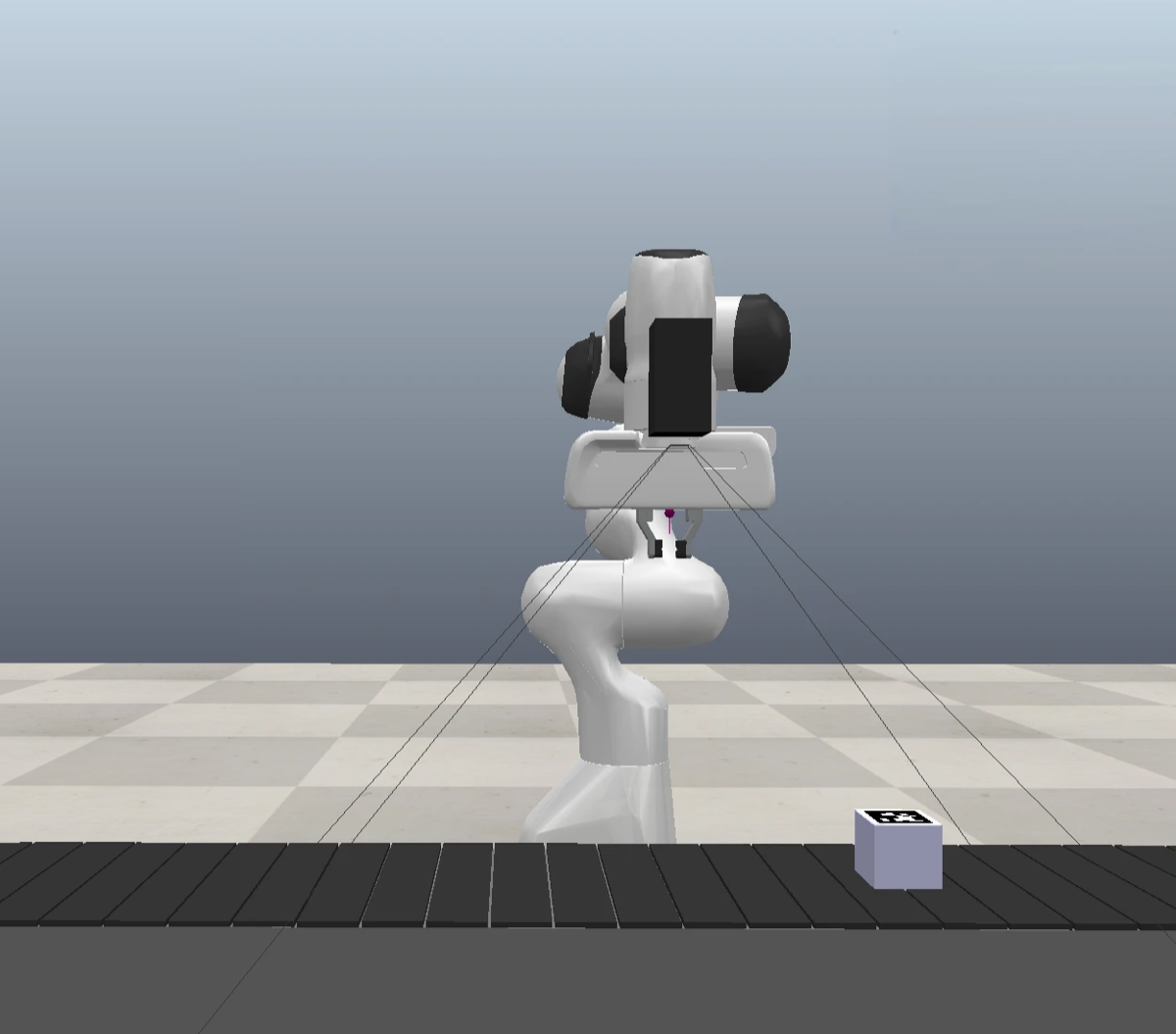}\\[3pt]
    \includegraphics{images/exp2.2_image.png} 
\end{subfigure}
\hfil
\begin{subfigure}{0.24\linewidth}
    \includegraphics{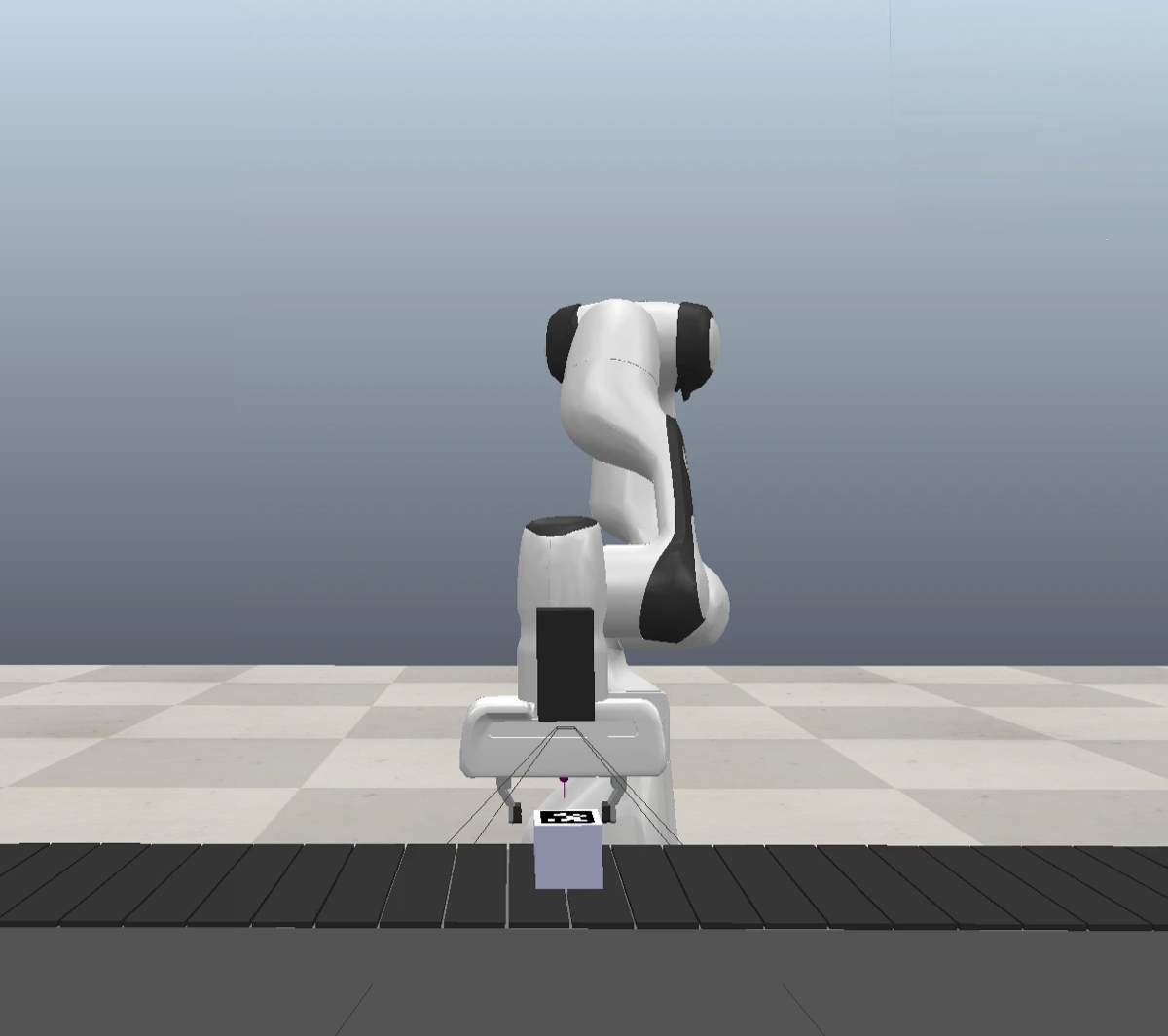}\\[3pt]
    \includegraphics{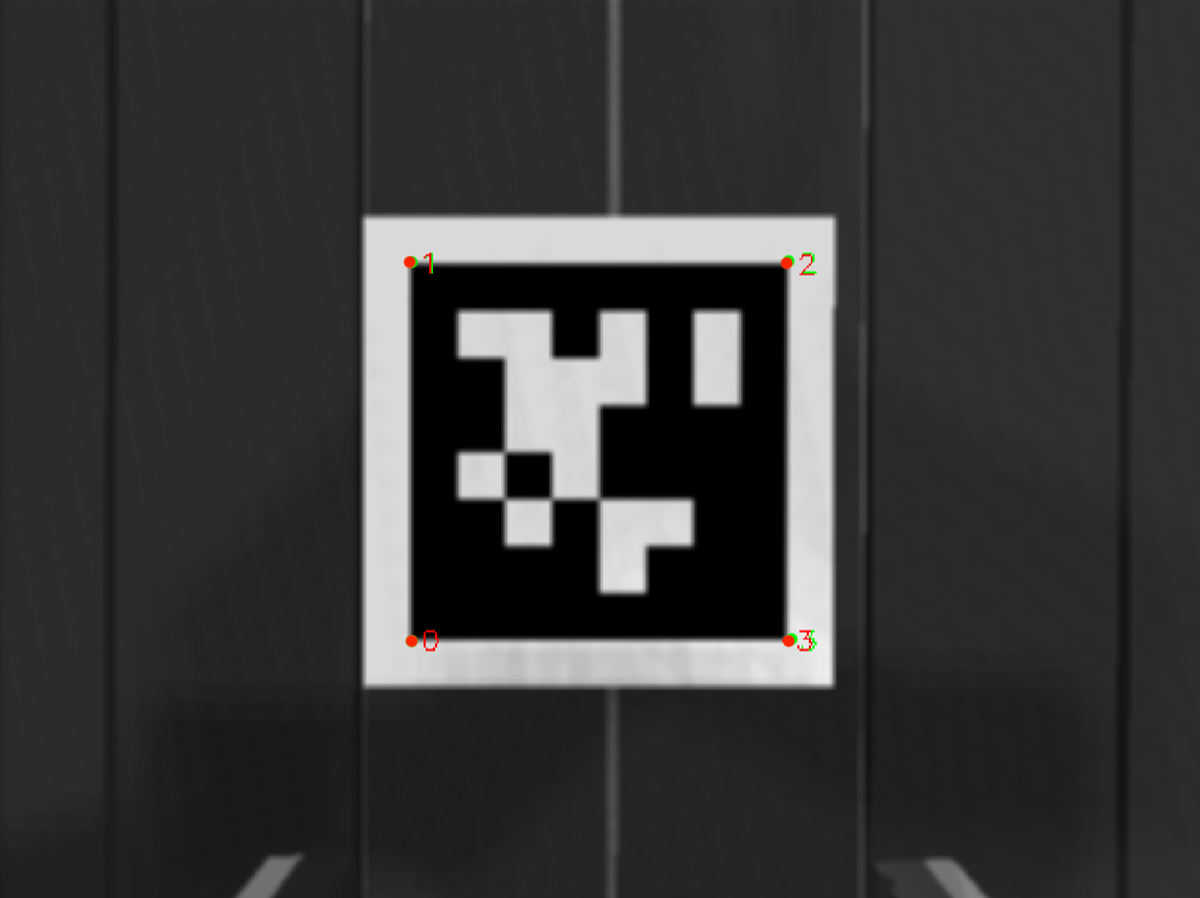}
\end{subfigure}
\hfil
\begin{subfigure}{0.24\linewidth}
    \includegraphics{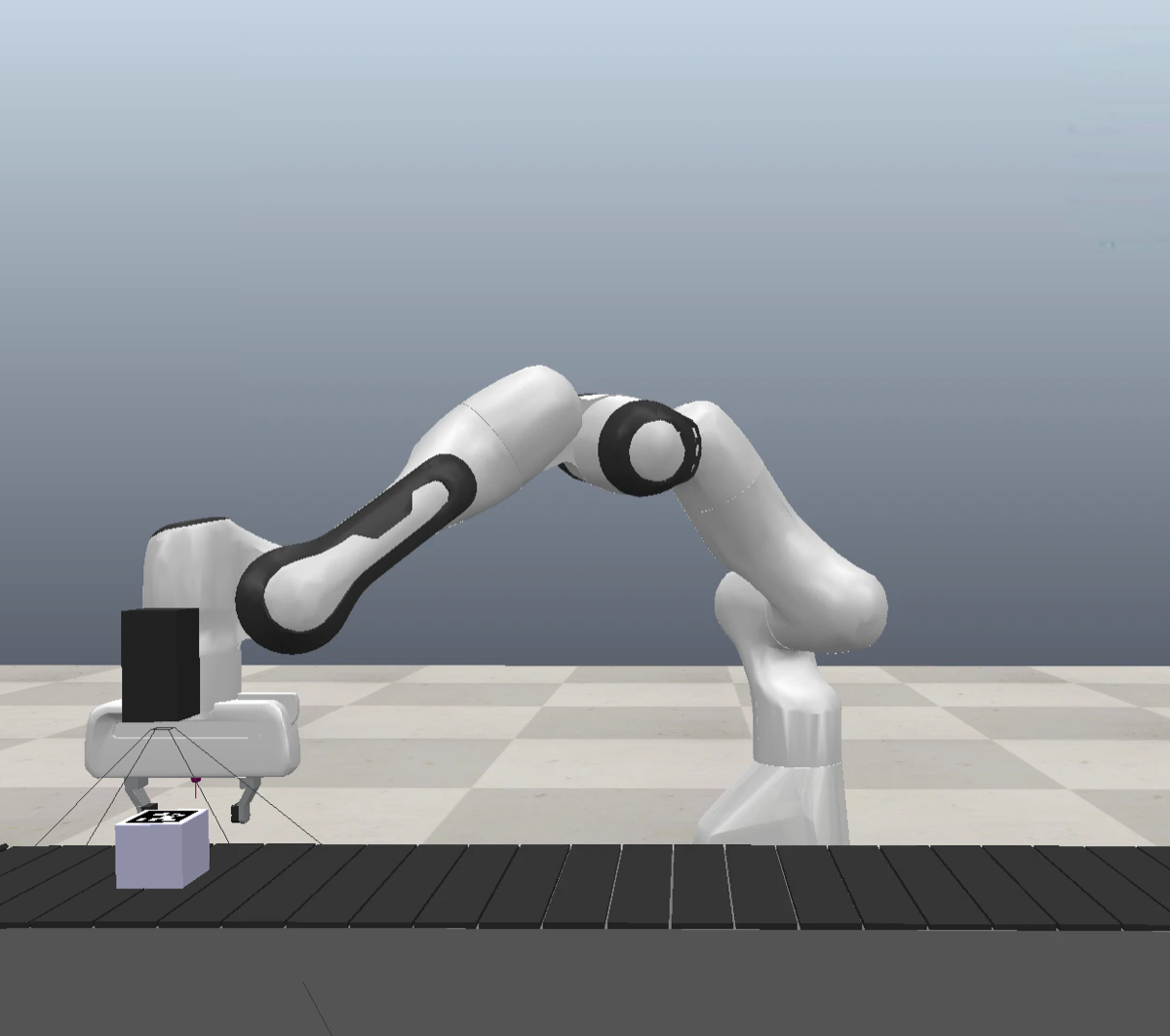}\\[3pt]
    \includegraphics{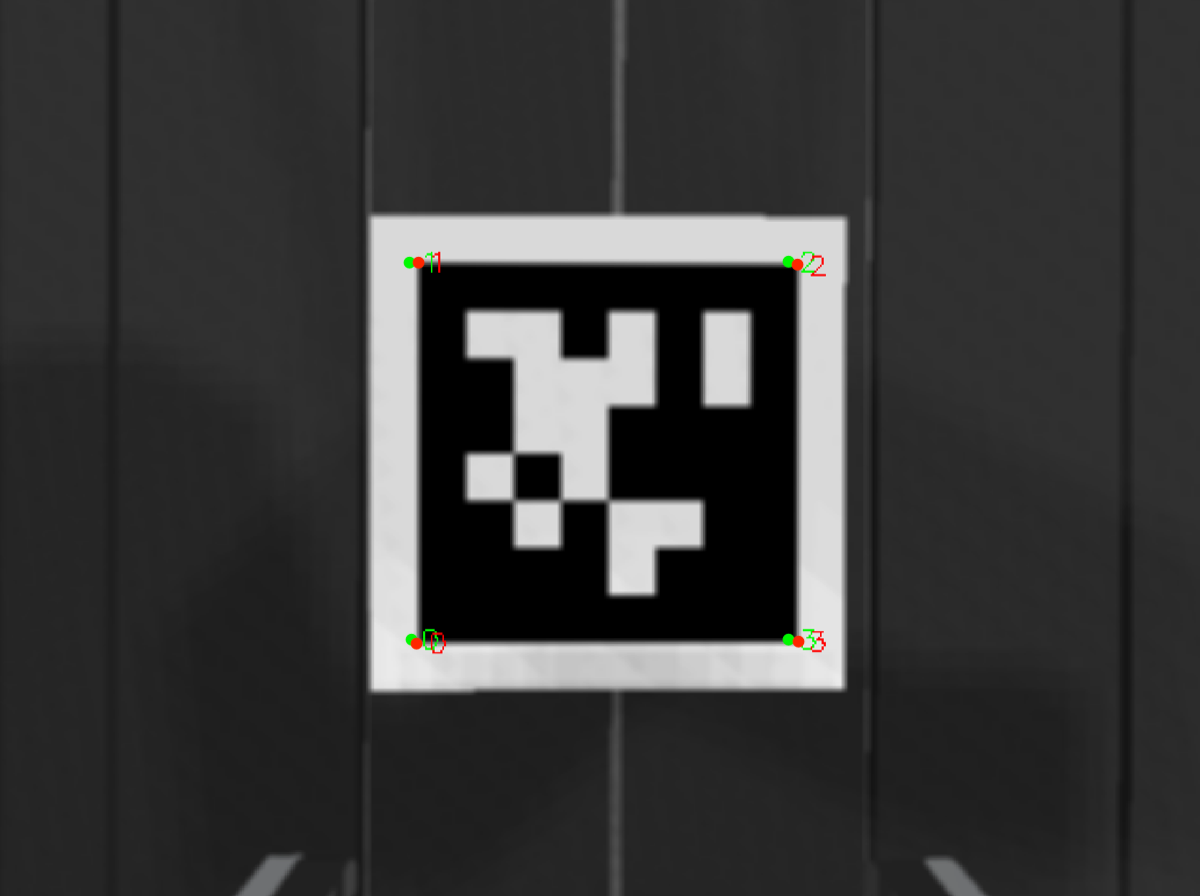}
\end{subfigure}
\caption{Snapshots of the \ac{ilvs} experiment with unseen initial position and demonstrated orientation: robot's external views (top) and camera images (bottom).}
\label{fig:adapt2}
\end{figure}

The second experiment of this set aims to evaluate the effectiveness of the approach in handling unseen conditions.
%
In particular, at the beginning of the experiment, the camera is oriented as in the demonstrations but has a substantial difference in position.
The large initial positional offset is well visible in the plot of Fig.~\ref{fig:exp2}, where the initial value of the visual features is far off from the demonstration.
Nevertheless, the visual features trajectories shown in Fig.~\ref{fig:exp2} (left) demonstrate that the robot manage to successfully achieve the \ac{vs} task, as the current value of the feature converges to their desired one, as also demonstrated in the dataset.
Similarly, target tracking performance can be evaluated also from the time evolution of the visual error presented in Fig.~\ref{fig:exp2} (right).
From this plot, one can evaluate that the visual error is kept to zero after a transient time, even while the box continues moving on the conveyor belt.
Four snapshots of this \ac{ilvs} experiment can be evaluated in Fig.~\ref{fig:adapt2}: the manipulator can reach the box and keep it tracking for all the experiments. 
The last two snapshots show how the robot manages to keep the box at the center of the image for the experiment, accommodating the motion induced by the conveyor belt.

\begin{figure}[t!]%
\centering%
 \begin{subfigure}[b]{0.5\textwidth}%
    \includegraphics[width=\columnwidth]{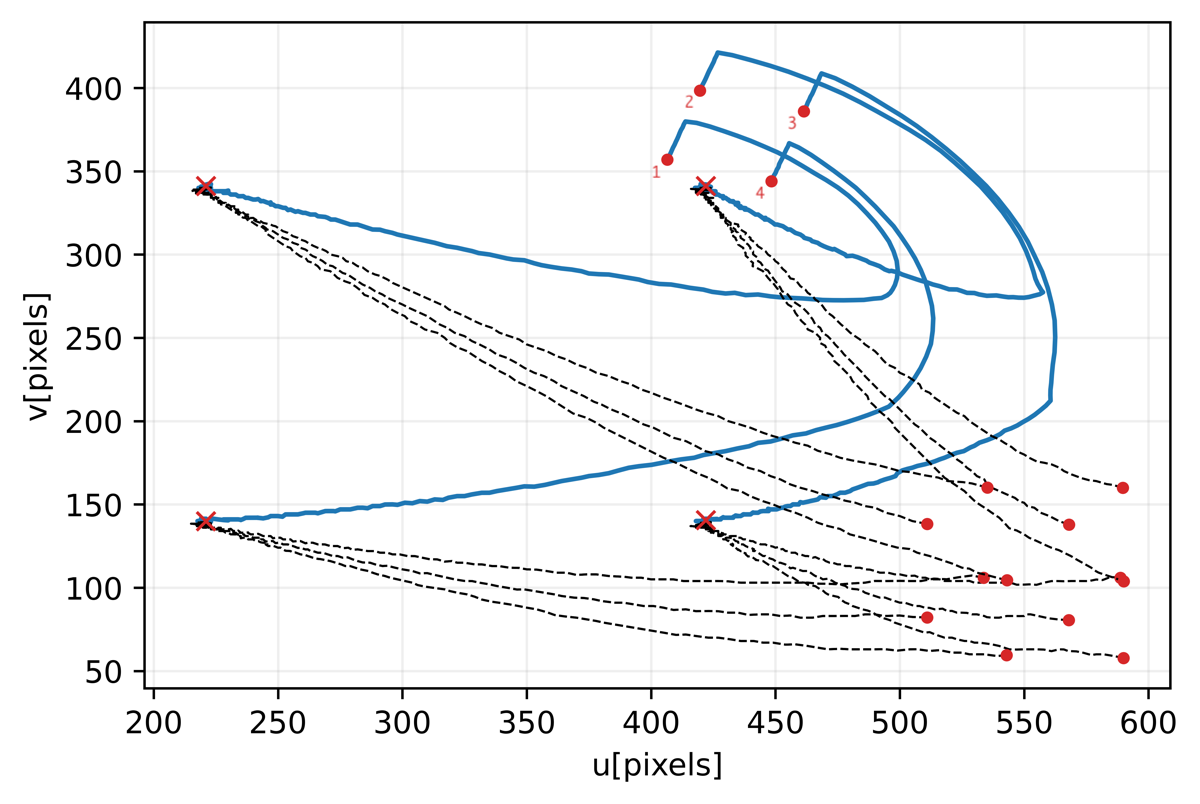} %
\end{subfigure}%
\hfill
\begin{subfigure}[b]{0.5\textwidth}%
     \includegraphics[width=\columnwidth]{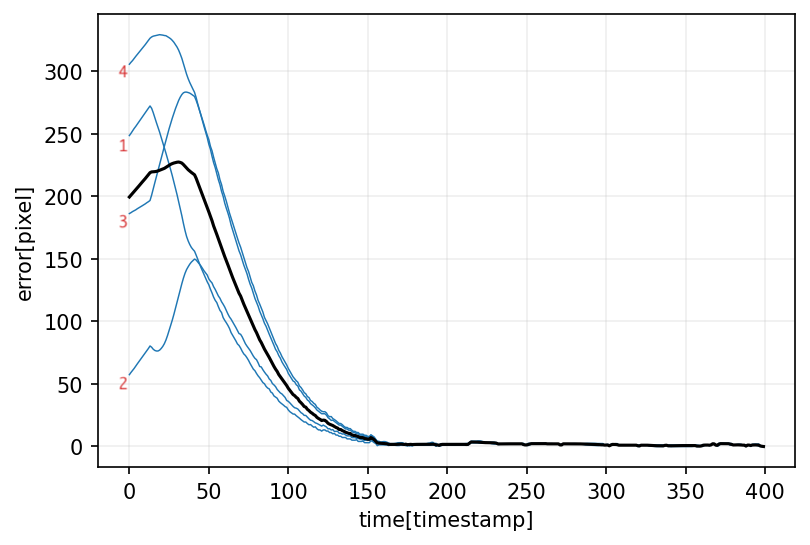}%
\end{subfigure}%
\caption{\ac{ilvs} experiment with unseen initial position and orientation: visual features trajectories~(left) and visual error~(right).}
\label{fig:exp3}%
\end{figure}

\begin{figure}[!t]
\centering
\setkeys{Gin}{width=\linewidth}
\begin{subfigure}{0.24\textwidth}
    \includegraphics[
  height=2.63cm
]{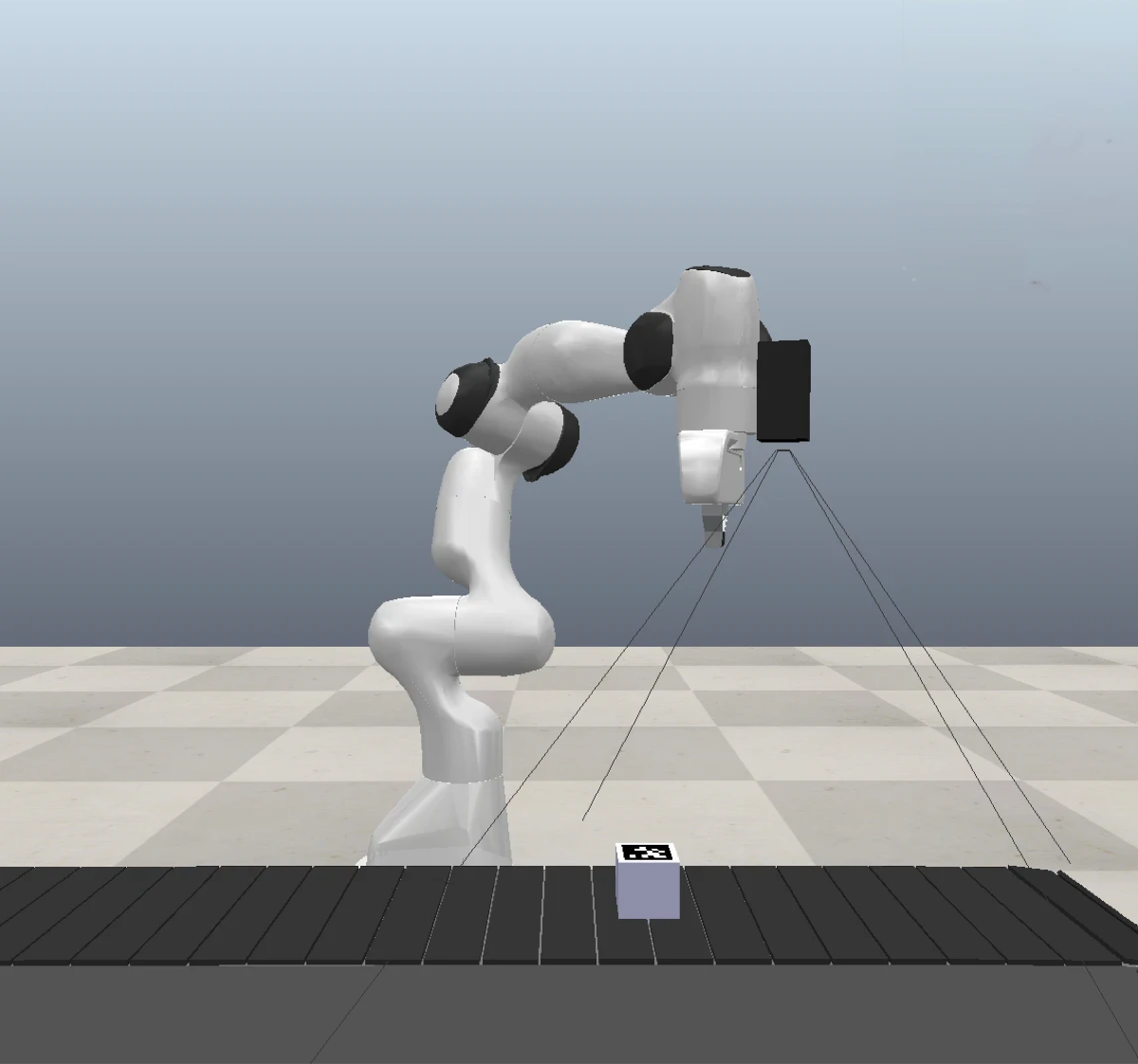}\\[3pt]
    \includegraphics{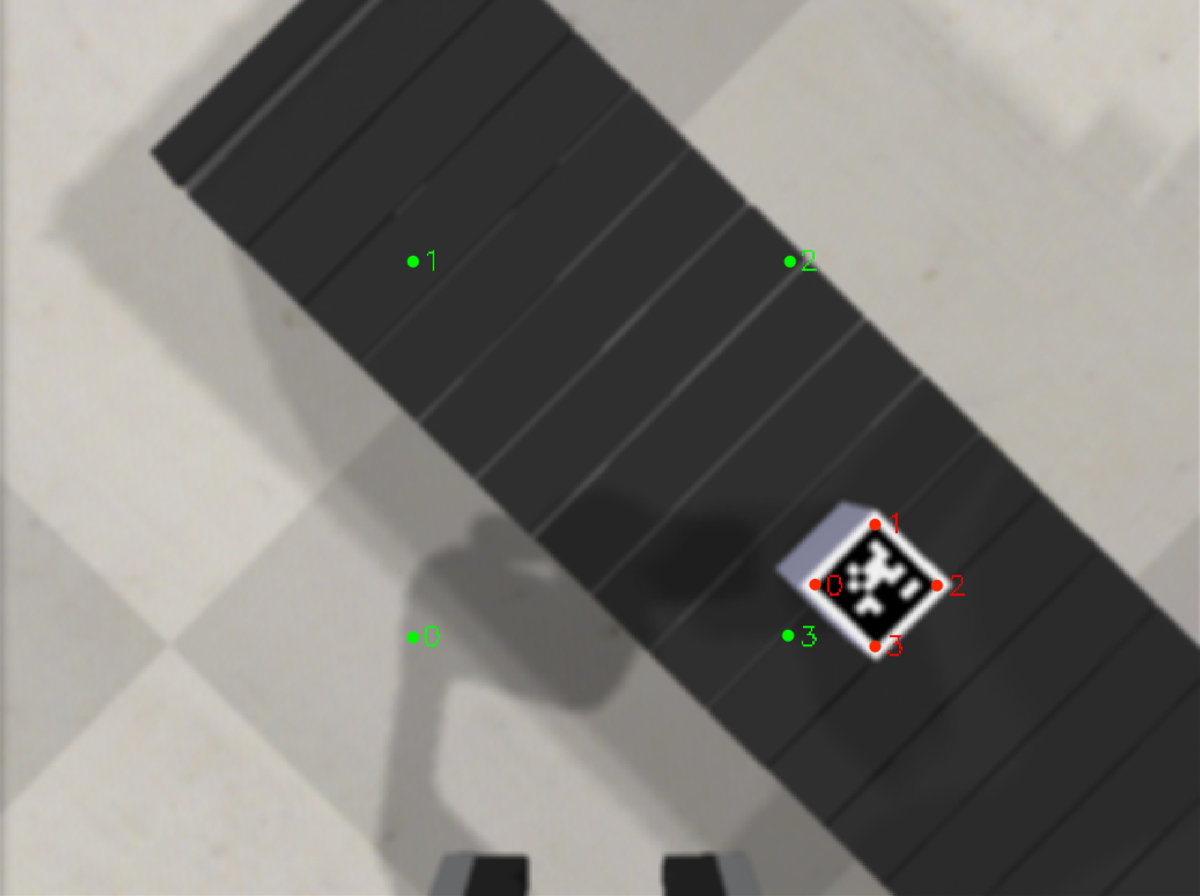}
\end{subfigure}
\hfil
\begin{subfigure}{0.24\linewidth}
    \includegraphics{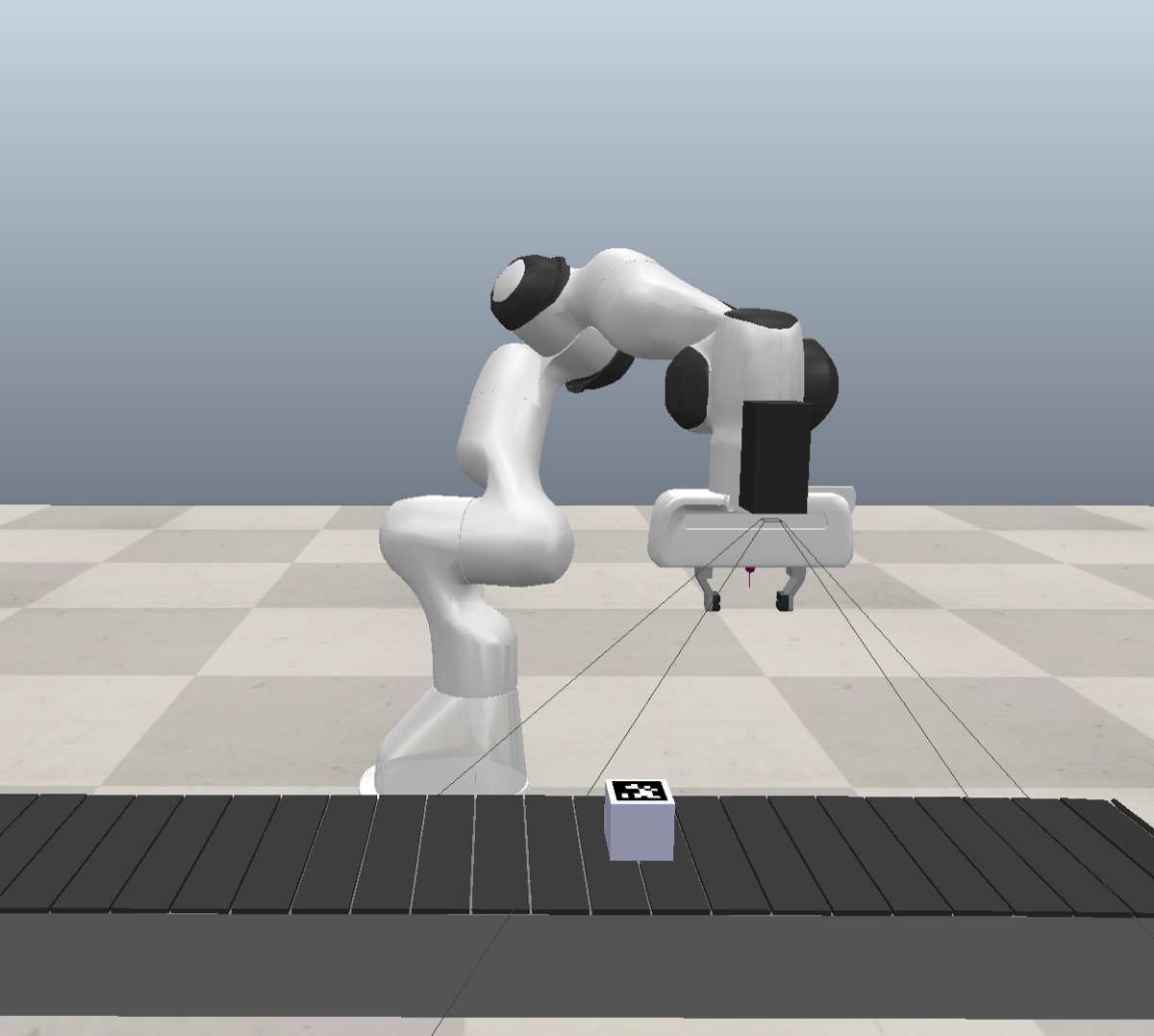}\\[3pt]
    \includegraphics{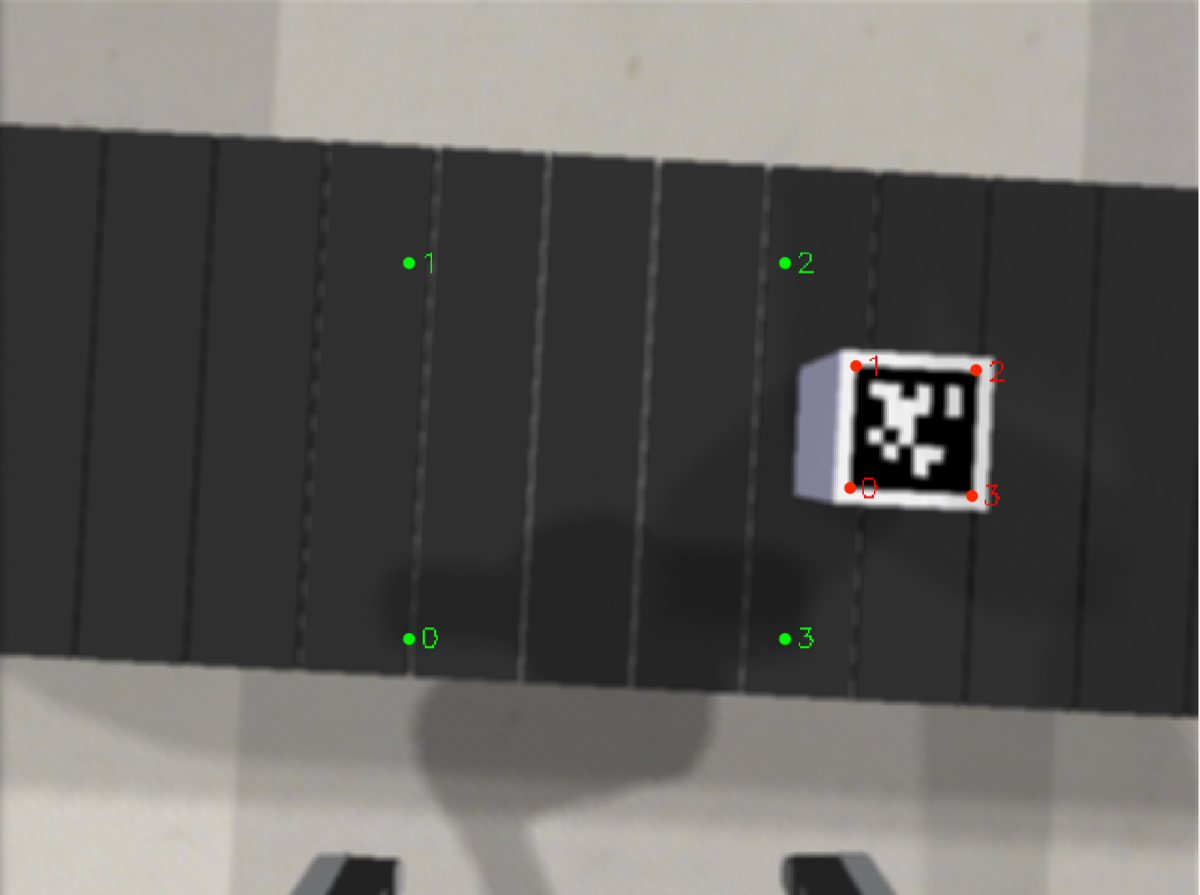} 
\end{subfigure}
\hfil
\begin{subfigure}{0.24\linewidth}
    \includegraphics{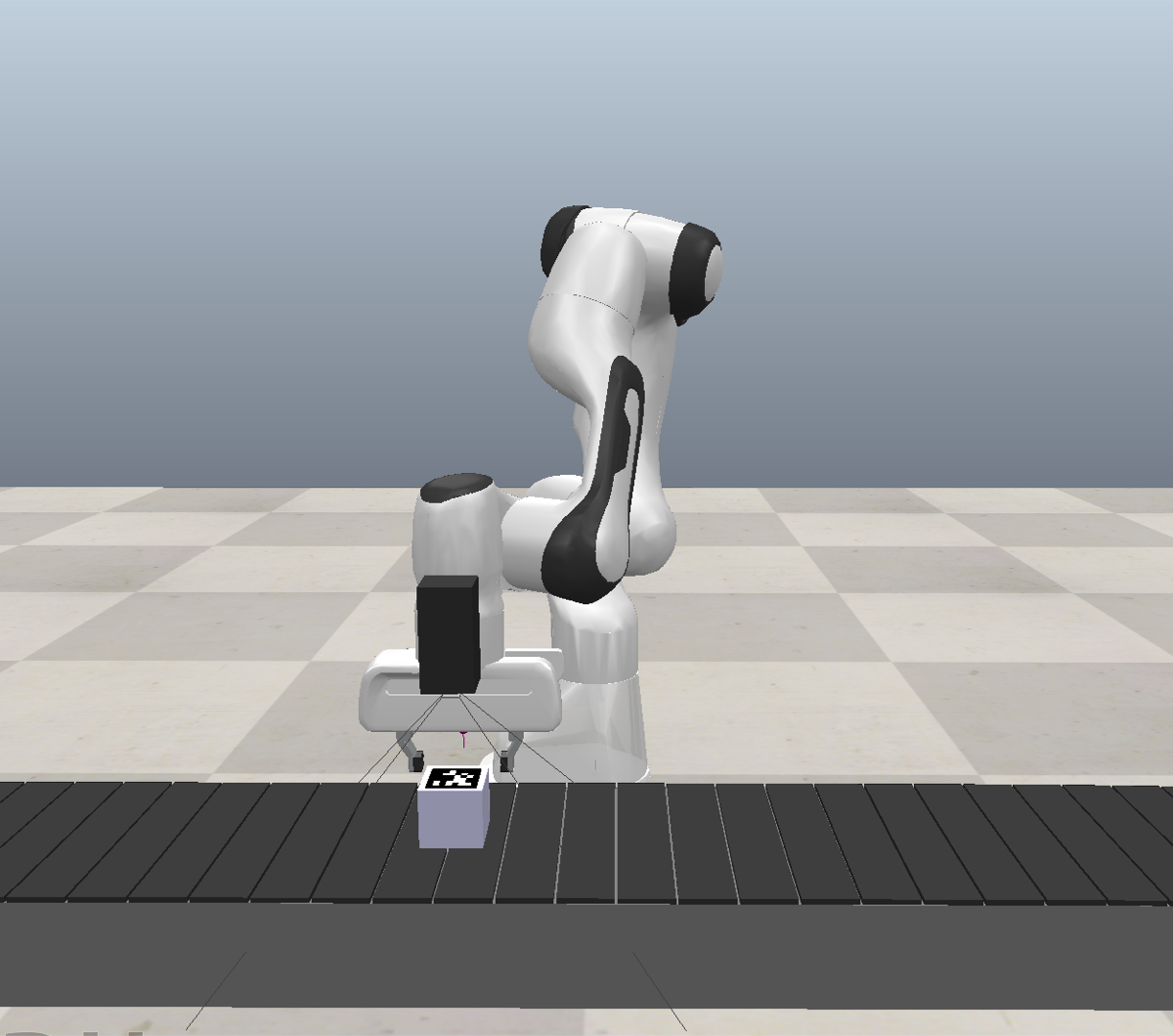}\\[3pt]
    \includegraphics{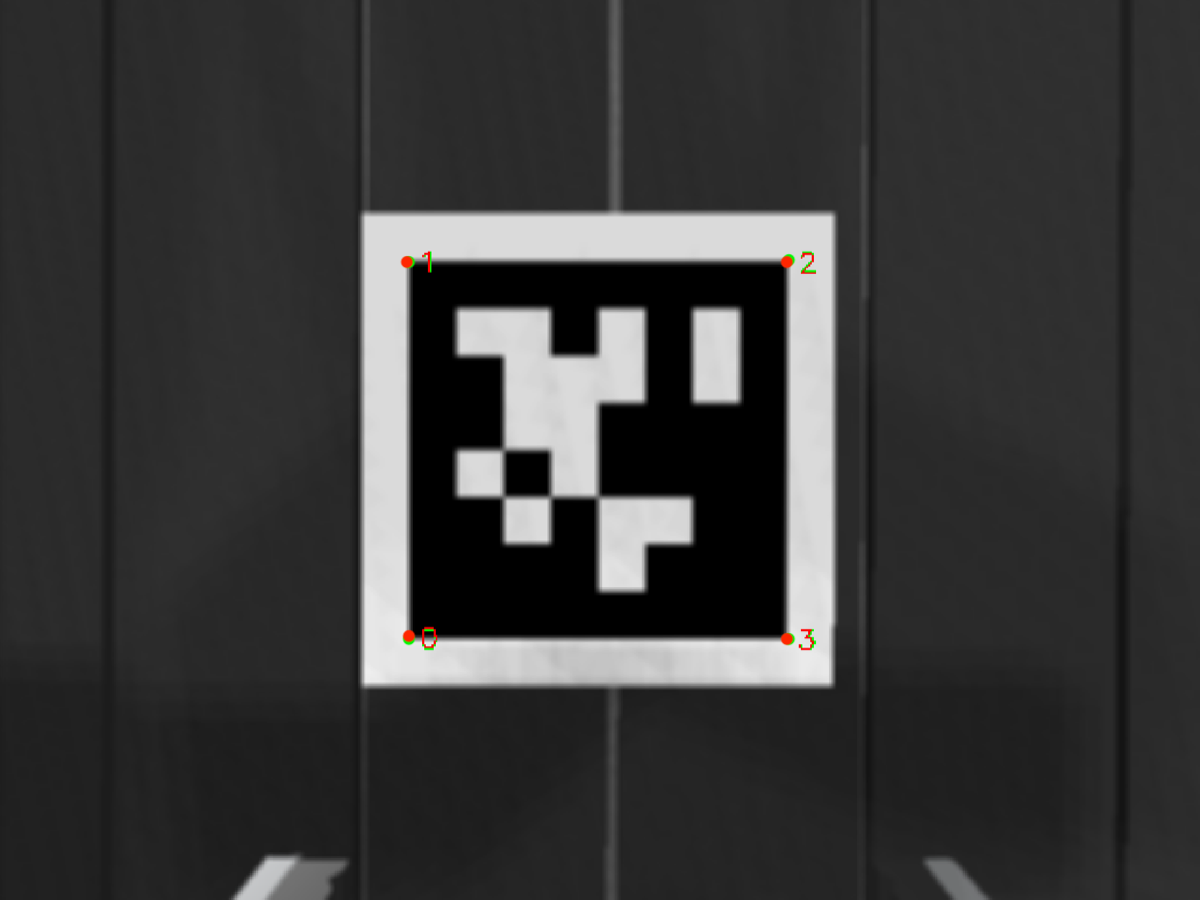}
    \end{subfigure}
    \begin{subfigure}{0.24\linewidth}
    \includegraphics{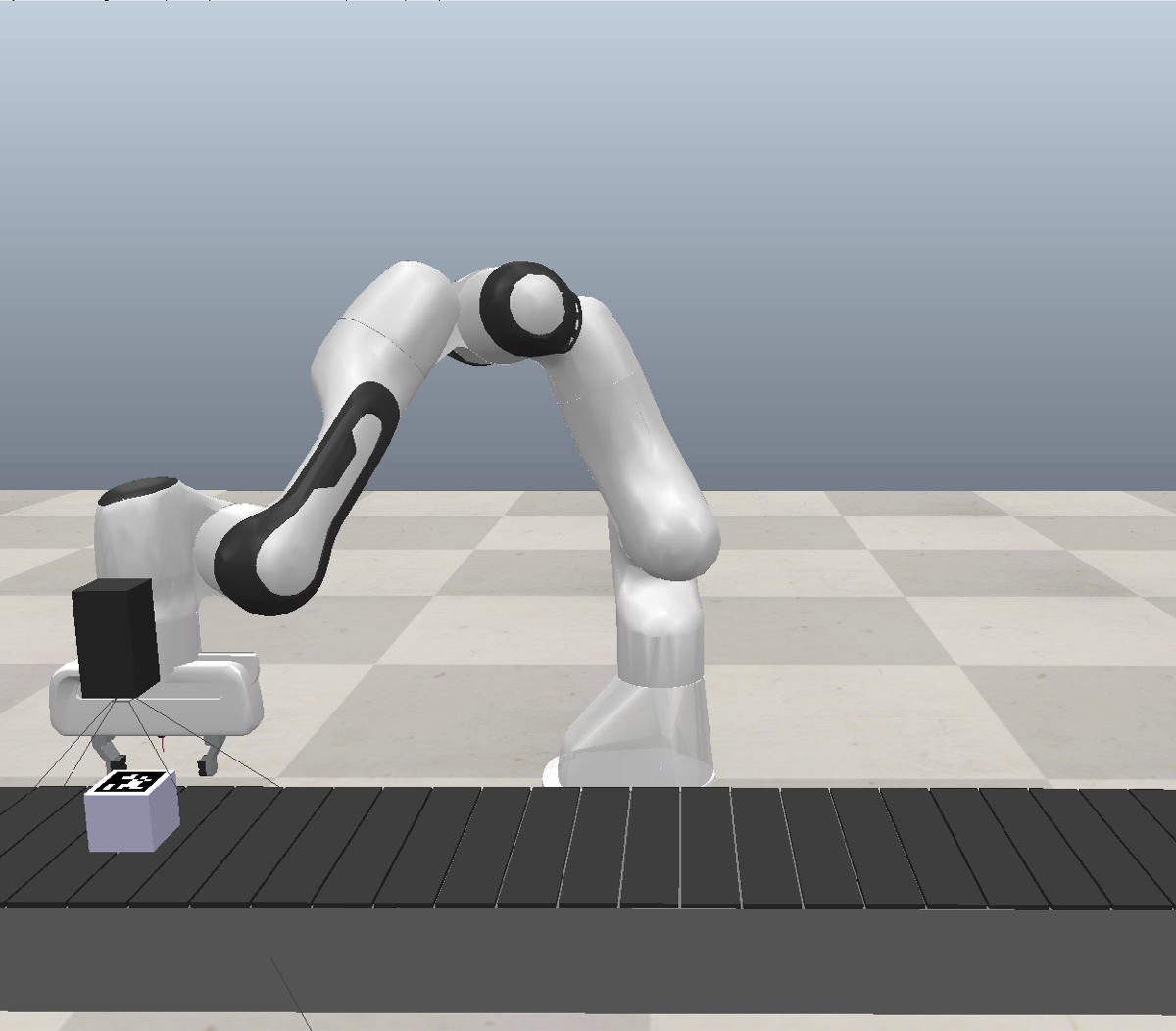}\\[3pt]
    \includegraphics{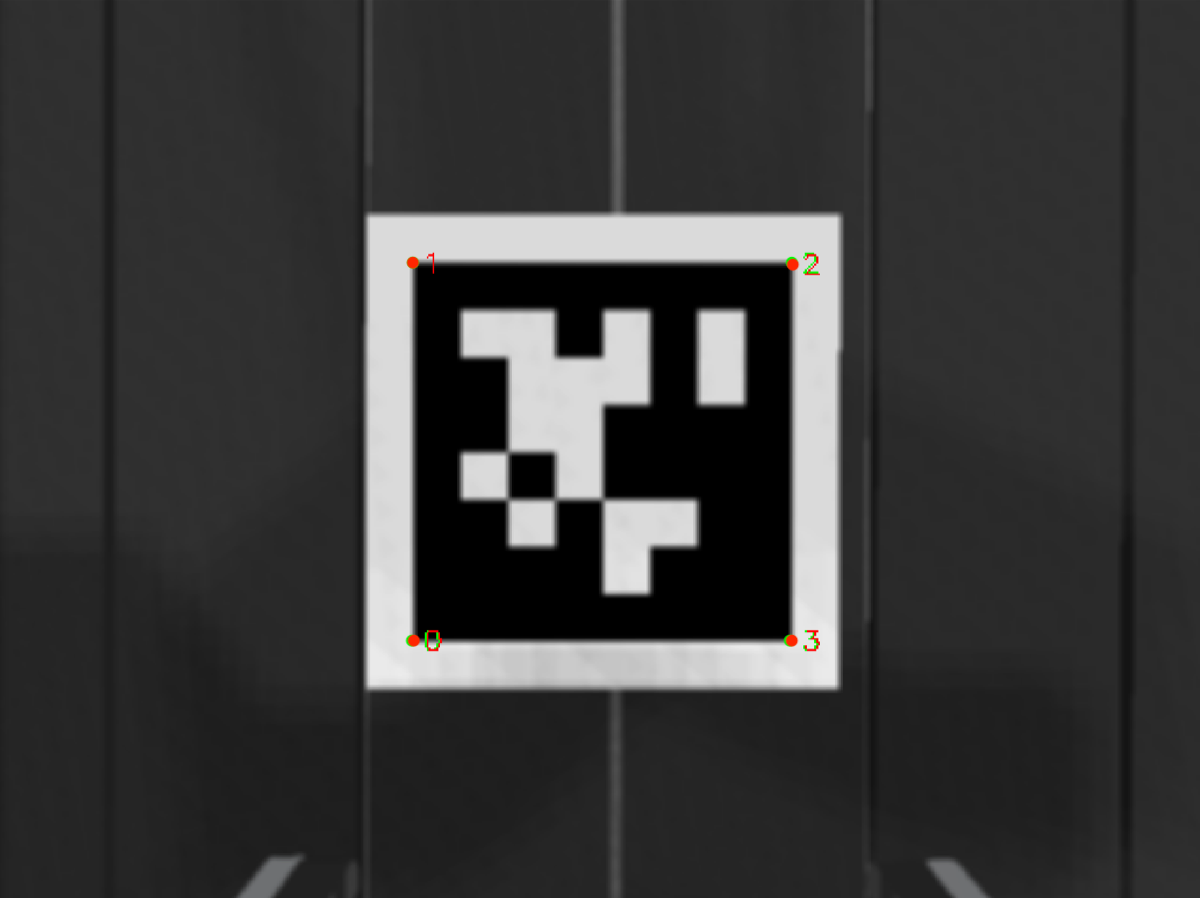}
\end{subfigure}
    \caption{Snapshots of the \ac{ilvs} experiment with unseen initial position and orientation: robot's external views (top) and camera images (bottom).} 
    \label{fig:adapt3}
\end{figure}

\begin{figure}[h!]%
    \centering%
    \includegraphics[width=0.32\columnwidth]{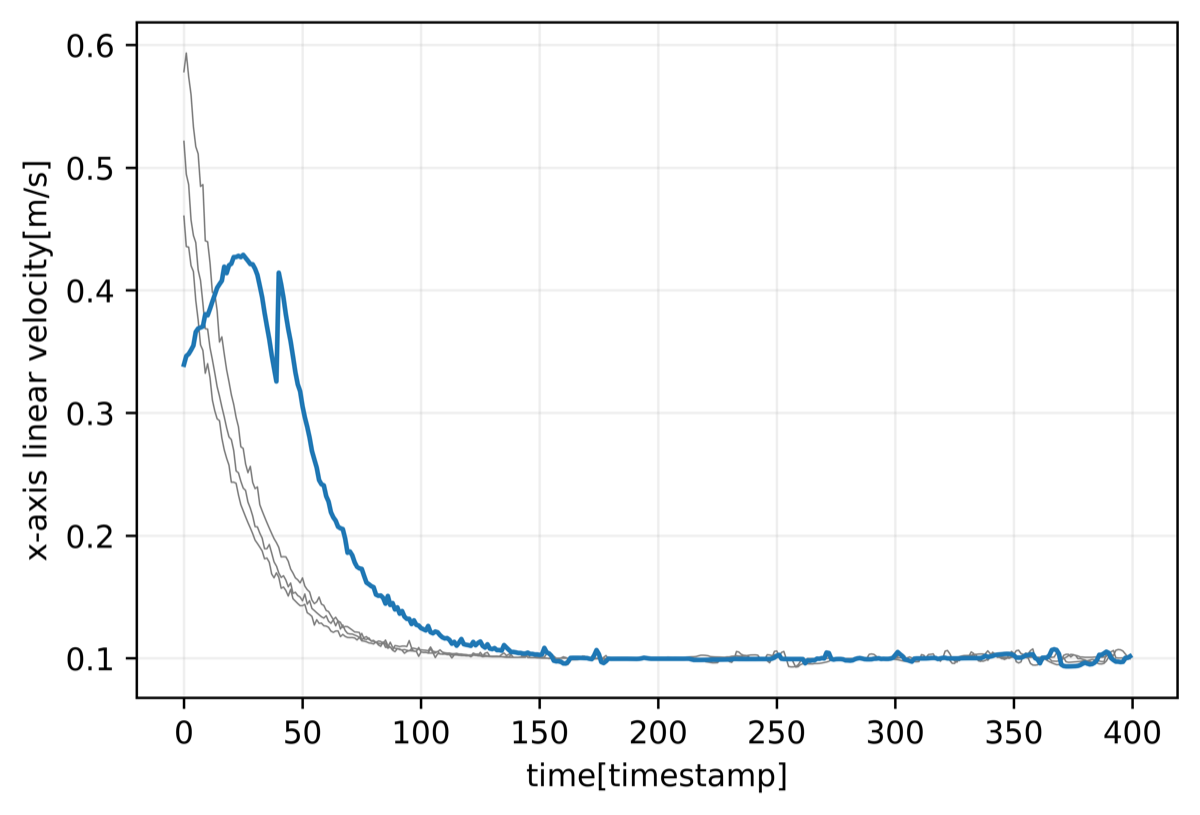} \hfill
    \includegraphics[width=0.32\columnwidth]{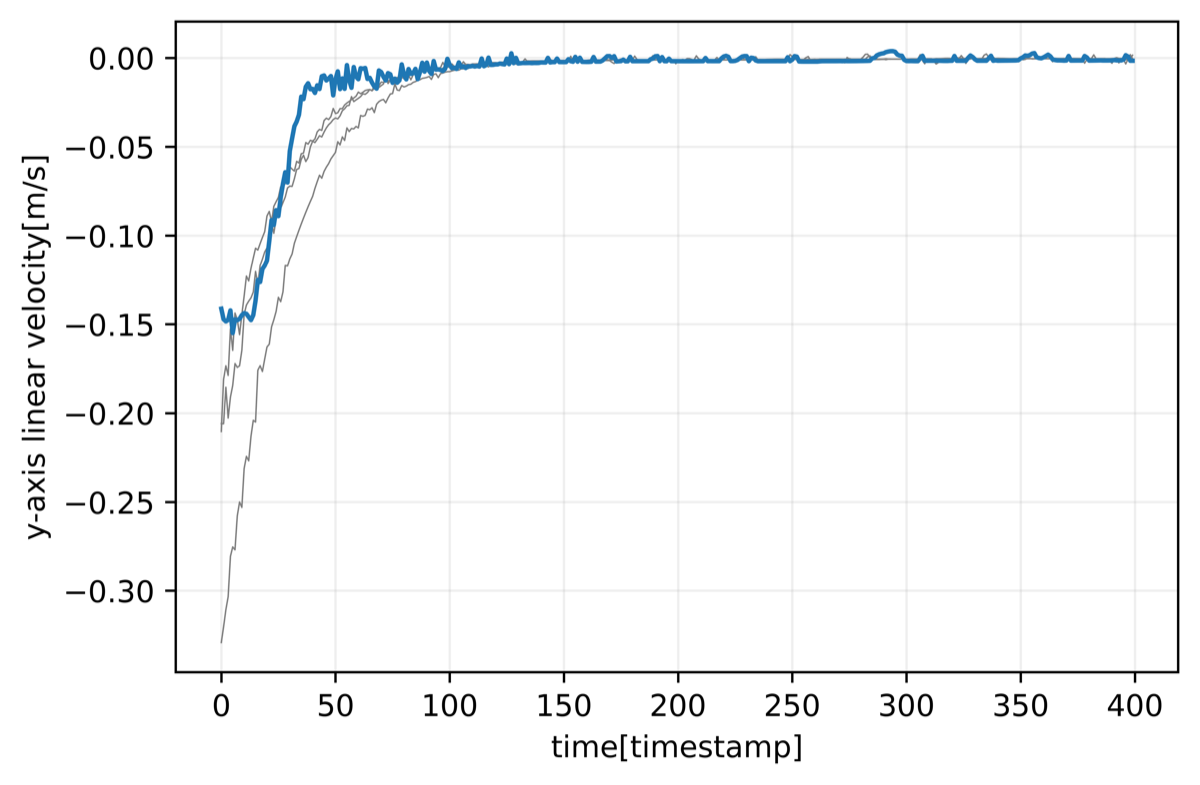} \hfill
    \includegraphics[width=0.31\columnwidth]{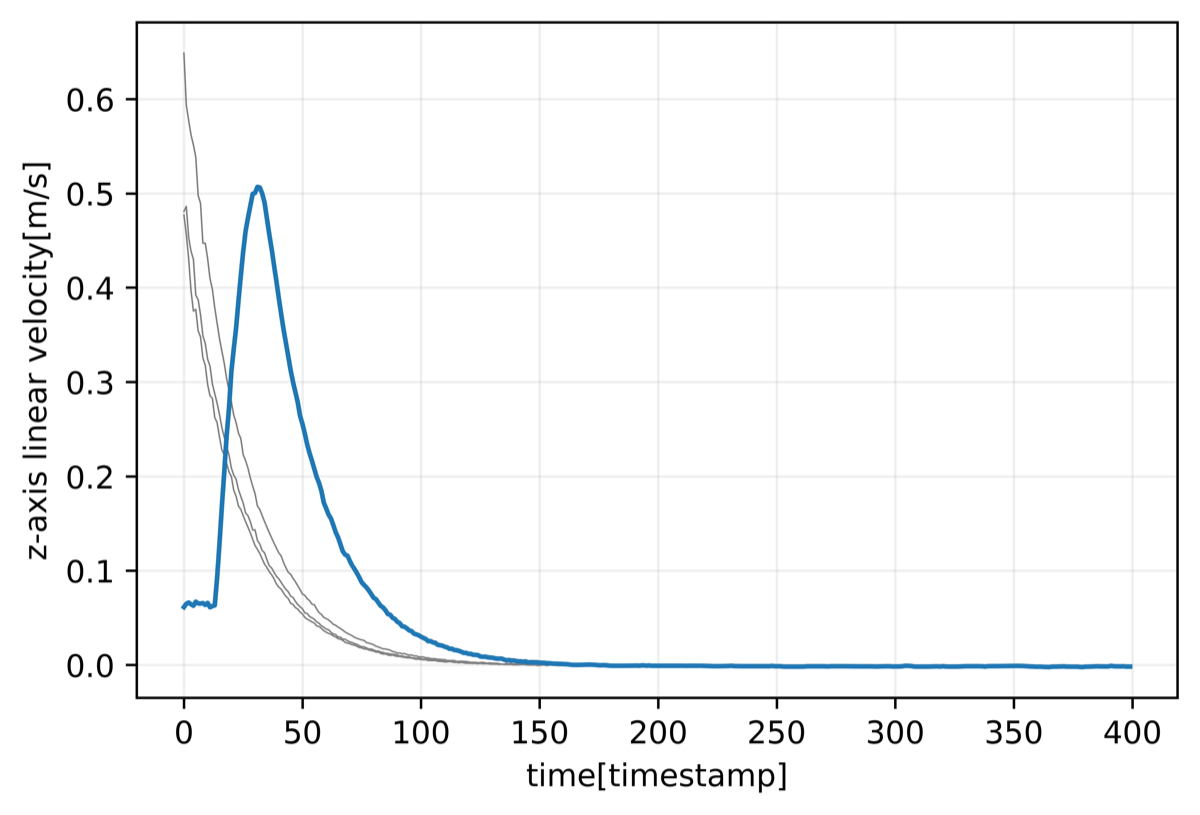} \hfill
    \includegraphics[width=0.32\columnwidth]{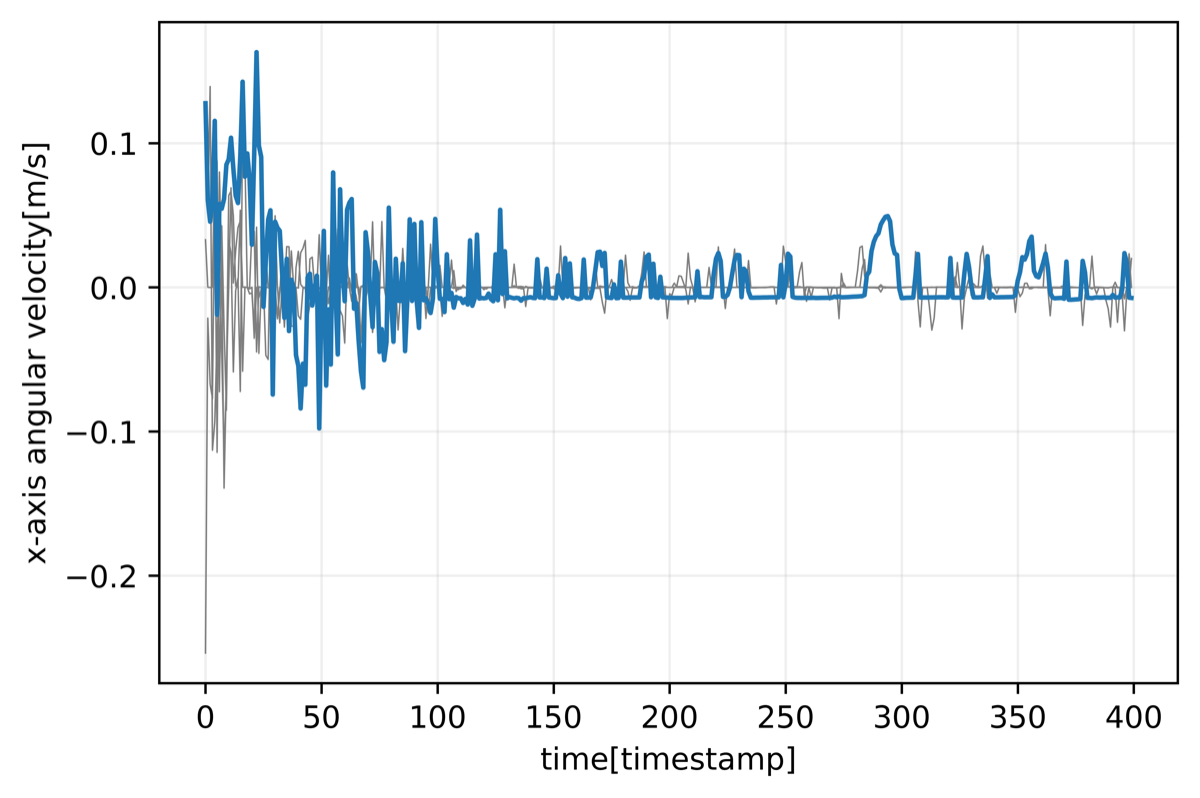} \hfill
    \includegraphics[width=0.32\columnwidth]{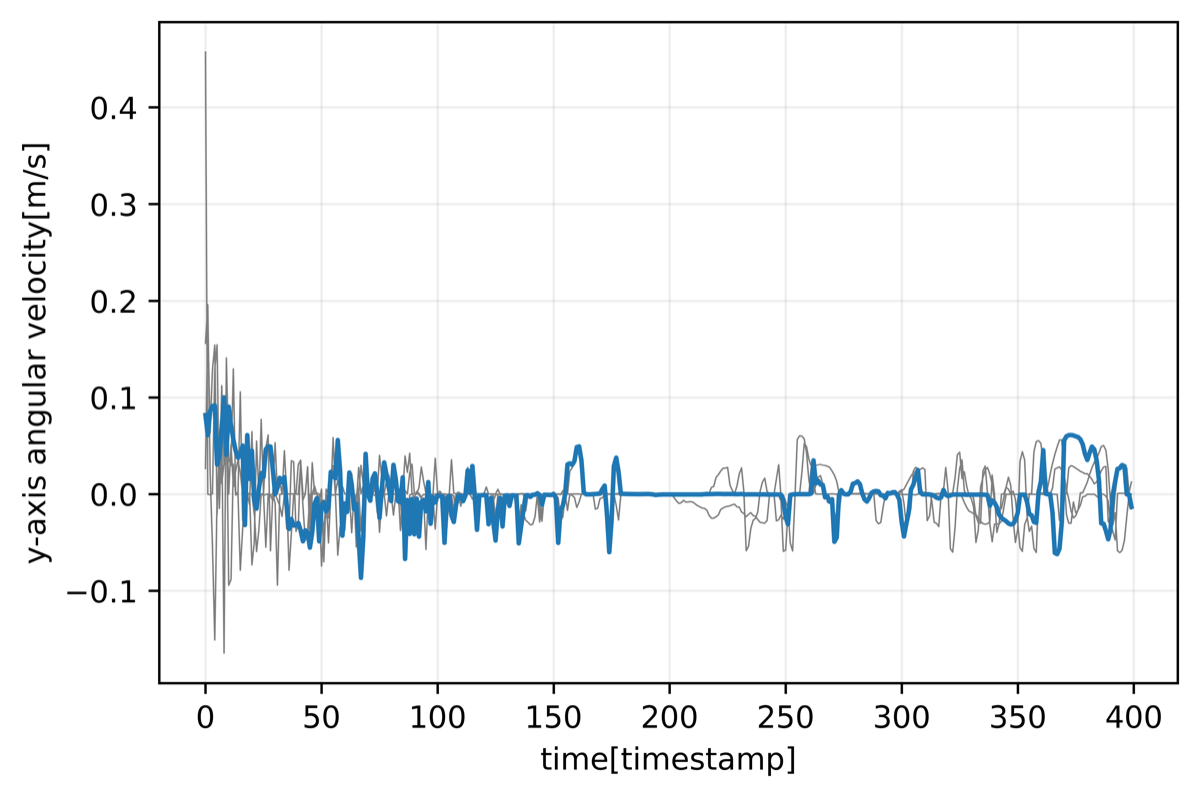} \hfill
    \includegraphics[width=0.31\columnwidth]{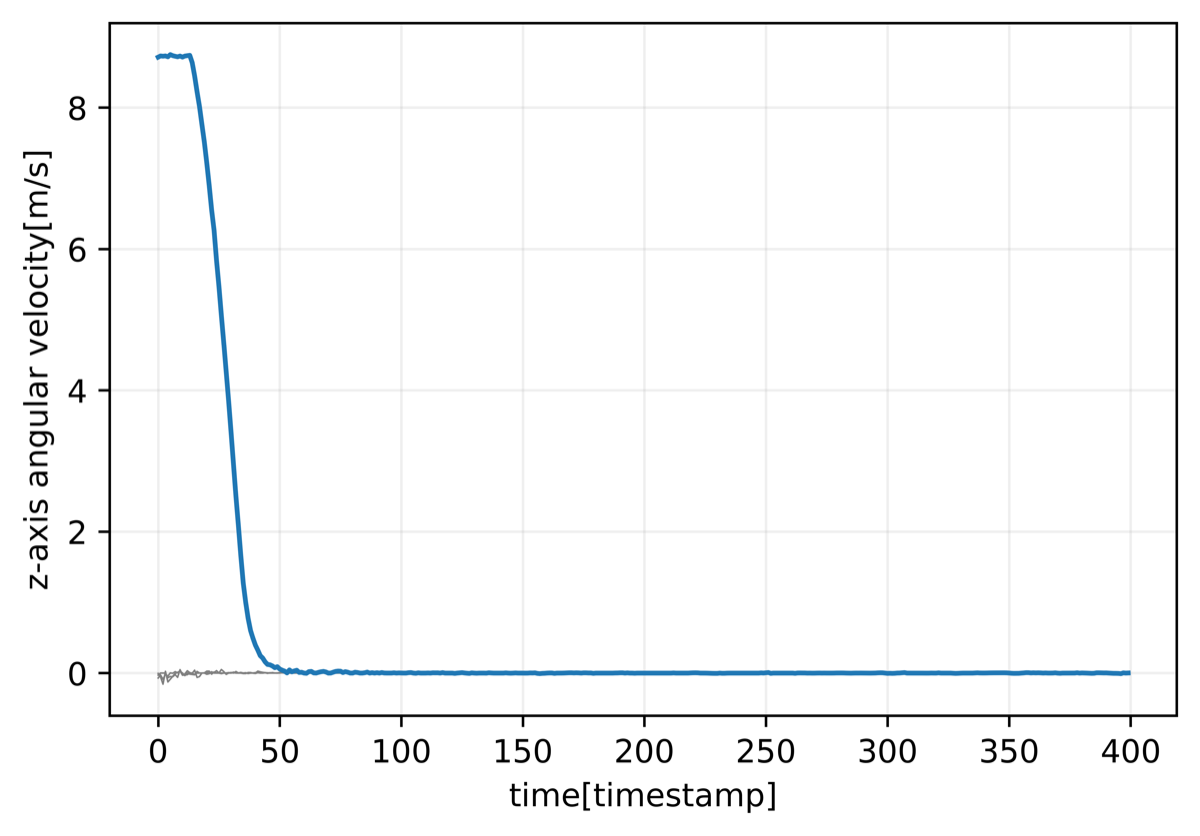}%
    \caption{Camera velocity during the ILVS experiment with unseen initial position and orientation: linear (top) and angular components (down).}
    \label{fig:vel3}
\end{figure}

The third experiment is meant to test at its greatest degree the handling of unseen initial conditions. 
As can be seen in Fig.~\ref{fig:exp3} (left) from the position of the features in the image plane, the end-effector of the manipulator at the beginning of the experiment has a pose that is not present in the training data. 
Nevertheless, the robot still manages to adjust its movement to successfully approach the moving target ensuring convergence, and once reached, it is able to track the target along its motion (see also the snapshots of Fig.~\ref{fig:adapt3}).
For this experiment, we also show in Fig.~\ref{fig:vel3} the plots of the camera velocities, as demonstrated (grey lines in the plots) and as executed by our method (in blue). 

For these three experiments, we provide a quantitative evaluation of the tracking performances. 
In particular, we considered the phase of the experiments that starts when the visual error is lower than $5$~pixels (cfr. Fig.~\ref{fig:exp1} (right), Fig.~\ref{fig:exp2} (right), and Fig.~\ref{fig:exp3} (right)).
For this portion of the experiments, 
the visual error is on average $1.795 \pm 0.984$~pixels, corresponding to $ 0.475 \pm 0.257$~mm of error in the camera position. 

Finally, we perform one last test in which we suddenly move the target object during the execution of the experiment. 
We observed the system's ability to adjust to such sudden and unexpected movements of the target object (tests were pursued with both low gain $\lambda$ = 2 and high gain $\lambda=10$ yielding satisfactory results in both cases).
The results of this experiment can be evaluated from the accompanying video.


\section{Discussion and conclusion}\label{sec:conclusion}

In this work, we have addressed some of the needs that arise from the introduction of friendly robots in domestic and industrial contexts where users are not necessarily experts.
In these situations, adaptability and easiness of use are must-haves for robots.
Therefore, we have proposed an imitation learning-based visual servoing framework for target tracking operations that avoids explicit programming, leveraging previous demonstrations of the desired behavior.
Our approach relies on the \ac{vs} paradigm and the \ac{ds}-based \ac{il} rationale. 
In particular, we take advantage of the imitation strategy to learn the compensation term required to achieve the visual tracking experiment.
Our approach permits us to realize the tracking without the specific implementation of an estimator or observer of the compensation term.
The framework has been evaluated with several simulations, which show the ability to handle unseen initial conditions.

As shown by the experiment in Fig.~\ref{fig:exp2} and Fig.~\ref{fig:adapt2}, the robot can converge to the visual target even starting relatively far from the initial value of the demonstrations. 
This out-of-domain generalization capability is a structural property of our approach that effectively combines a stable component (from standard \ac{vs}) and a learned one in the closed-loop control law~\eqref{eq:ours}.
Indeed, the standard \ac{vs} component always drives the robot close to the target, i.e., in the training data domain, where the learning of the compensation term is put in an ideal condition to work.
Stronger generalization capabilities (e.g., to handle the doubled velocity of the conveyor belt seen during the demonstrations) would require re-training our compensation term.
The stability of the proposed controller has not been formally investigated (for instance, using tools from the Lyapunov theory).
However, in the conducted experiments, the robot was always able to reach the target with sub-millimeter precision. 
Moreover, we also tested the robustness to disturbances like changes in the object position on the conveyor belt. 
The fact that the controller behaved as expected in several practical cases suggests that it should have some (local) stability property. 
However, a formal stability proof is left as future work.
%
%
Another interesting line for future development is the test of our framework with velocities of the object that are different from the one seen during the demonstrations. Indeed in our current study, the velocity of the object during the validation experiment is the same as the one used during the collection of the demonstrations.
Finally, we plan to test our approach with real experiments;
to this end, further development will be required to handle the noise in the input data (typical of real-life applications).



\bibliographystyle{plain} 
\bibliography{biblio}

\end{document}